\documentclass{article}

\usepackage{microtype}
\usepackage{graphicx}
\usepackage{booktabs} % for professional tables

\usepackage{bm}

\usepackage{amsmath}
\usepackage{amsfonts}
\usepackage{amssymb}
\usepackage{graphicx}
\usepackage{amsmath}
\usepackage{amsfonts}
\usepackage{amssymb}
\usepackage{graphicx}
\usepackage{caption}
\usepackage{wrapfig}
\usepackage{nicefrac}
\usepackage{tikz}
\usepackage{pgfplots}
\usepackage{subcaption}

\DeclareMathOperator{\var}{VaR}
\DeclareMathOperator{\alphavar}{VaR_\alpha}
\DeclareMathOperator{\alphacvar}{CVaR_\alpha}
\DeclareMathOperator{\cvar}{CVaR}
\newcommand{\Ex}{\mathbb{E}}

\newcommand{\states}{\mathcal{S}}
\newcommand{\actions}{\mathcal{A}}
\newcommand{\Real}{\mathbb{R}}

\newcommand{\wb}{\bm{w}}

\renewcommand{\Pr}{\mathbb{P}}
\renewcommand{\rho}{v}
\usepackage{color}
\usepackage{xcolor}

\usepackage{hyperref}

\usepackage{color}

\usepackage[accepted]{icml2021}

\icmltitlerunning{Policy Gradient Bayesian Robust Optimization for Imitation Learning}

\begin{document}

\twocolumn[
%\icmltitle{Bayesian Robust Policy Optimization Under Reward Function Uncertainty}
\icmltitle{Policy Gradient Bayesian Robust Optimization for Imitation Learning}



\icmlsetsymbol{equal}{*}

\begin{icmlauthorlist}
\icmlauthor{Zaynah Javed}{equal,ucb}
\icmlauthor{Daniel S. Brown}{equal,ucb}
\icmlauthor{Satvik Sharma}{ucb}
\icmlauthor{Jerry Zhu}{ucb}
\icmlauthor{Ashwin Balakrishna}{ucb}
\icmlauthor{Marek Petrik}{unh}
\icmlauthor{Anca D. Dragan}{ucb}
\icmlauthor{Ken Goldberg}{ucb}
% \icmlauthor{Bauiu C.~Yyyy}{equal,to,goo}
% \icmlauthor{Cieua Vvvvv}{goo}
% \icmlauthor{Iaesut Saoeu}{ed}
% \icmlauthor{Fiuea Rrrr}{to}
% \icmlauthor{Tateu H.~Yasehe}{ed,to,goo}
\end{icmlauthorlist}

\icmlaffiliation{ucb}{EECS Department, University of California, Berkeley}
\icmlaffiliation{unh}{CS Department, University of New Hampshire}

\icmlcorrespondingauthor{Zaynah Javed}{zjaved@berkeley.edu}
\icmlcorrespondingauthor{Daniel Brown}{dsbrown@berkeley.edu}

\icmlkeywords{Machine Learning, ICML}

\vskip 0.3in
]


% \icmlEqualContribution

\printAffiliationsAndNotice{}  % leave blank if no need to mention equal contribution

\begin{abstract}
The difficulty in specifying rewards for many real-world problems has led to an increased focus on learning rewards from human feedback, such as demonstrations. However, there are often many different reward functions that explain the human feedback, leaving agents with \emph{uncertainty} over what the true reward function is. While most policy optimization approaches handle this uncertainty by optimizing for expected performance, many applications demand risk-averse behavior. 
We derive a novel policy gradient-style robust optimization approach, PG-BROIL, that optimizes a soft-robust objective that balances expected performance and risk. To the best of our knowledge, PG-BROIL is the first policy optimization algorithm robust to a distribution of reward hypotheses which can scale to continuous MDPs.
Results suggest that PG-BROIL can produce a family of behaviors ranging from risk-neutral to risk-averse and outperforms state-of-the-art imitation learning algorithms when learning from ambiguous demonstrations by hedging against uncertainty, rather than seeking to uniquely identify the demonstrator's reward function. 
\end{abstract}

\section{Introduction}
We consider the following question: \textit{How should an intelligent agent act if it has epistemic uncertainty over its objective function?} In the fields of reinforcement learning (RL) and optimal control, researchers and practitioners typically assume a known reward or cost function, which is then optimized to obtain a policy. However, even in settings where the reward function is specified, it is usually only a best approximation of the objective function that a human thinks will lead to desirable behavior. Furthermore, human-designed reward functions are also often augmented with human feedback. This may also result in reward uncertainty since human feedback, be it in the form of policy shaping~\cite{griffith2013policy}, reward shaping~\cite{knox2012reinforcement}, or a hand-designed reward function~\cite{hadfield2017inverse,ratner2018simplifying}, can fail to perfectly disambiguate the human's intent true~\cite{amodei2016concrete}.

Reward function ambiguity is also a key problem in imitation learning \cite{hussein2017imitation,osa2018algorithmic}, in which an agent seeks to learn a policy from demonstrations without access to the reward function that motivated the demonstrations. While many imitation learning approaches either sidestep learning a reward function and directly seek to imitate demonstrations \cite{pomerleau1991efficient,torabi2018behavioral} or take a maximum likelihood~\cite{choi2011map,brown2019extrapolating} or maximum entropy approach to learning a reward function \cite{ziebart2008maximum,fu2017learning}, we believe that an imitation learning agent should explicitly reason about uncertainty over the true reward function to avoid misalignment with the demonstrator's objectives~\cite{hadfield2017inverse,brown2020safe}. Bayesian inverse reinforcement learning (IRL) methods \cite{ramachandran2007bayesian} seek a posterior distribution over likely reward functions given demonstrations, but often perform policy optimization using the expected reward function or MAP reward function \cite{ramachandran2007bayesian,choi2011map,ratner2018simplifying,brown2020safe}. However, in many real world settings such as robotics, finance, and healthcare, we desire a policy which is robust to uncertainty over the true reward function. 

Prior work on risk-averse and robust policy optimization in reinforcement learning has mainly focused on robustness to uncertainty over the true dynamics of the environment, but assumes a known reward function \cite{garcia2015comprehensive,tamar2015optimizing,pmlr-v100-tang20a,derman2018soft,lobo2020soft, recovery-rl}. Some work addresses robust policy optimization under reward function uncertainty by taking a maxmin approach and optimizing a policy that is robust under the worst-case reward function \cite{syed2008apprenticeship,Regan2009,hadfield2017inverse,huang2018learning}. However, these approaches are limited to tabular domains, and maxmin approaches have been shown to sometimes lead to incorrect and overly pessimistic policy evaluations~\cite{brown2018efficient}.
% However, existing work either assumes a discrete MDP with known transition dynamics or access to an MDP solver that can be called hundreds of times, effectively limiting these approaches to tabular domains. Furthermore, optimizing only for the worst-case reward function can lead to overly pessimistic policies \cite{brown2018efficient,hadfield2017inverse}.
As an alternative to maxmin approaches, recent work~\cite{brown2020bayesian} proposed a linear programming approach, BROIL: Bayesian Robust Optimization for Imitation Learning, that balances risk-aversion (in terms of Conditional Value at Risk~\cite{rockafellar2000optimization}) and expected performance. This approach supports a family of solutions depending on the risk-sensitivity of the application domain. However, as their approach is built on linear programming, it cannot be applied in MDPs with continuous state and action spaces and unknown dynamics.

In this work, we introduce a novel policy optimization approach that enables varying degrees of risk-sensitivity by reasoning about reward uncertainity while scaling to continuous MDPs with unknown dynamics. As in~\citet{brown2020bayesian}, we present an approach which reasons simultaneously about risk-aversion (in terms of Conditional Value at Risk~\cite{rockafellar2000optimization}) and expected performance and balances the two. However, to enable such reasoning in continuous spaces, we make a key observation: the Conditional Value at Risk objective supports efficient computation of an approximate subgradient, which can then be used in a policy gradient method. This makes it possible to use any policy gradient algorithm, such as TRPO~\cite{trpo} or PPO~\cite{schulman2017proximal} to learn policies which are robust to reward uncertainity, resulting in an efficient and scalable algorithm. To the best of our knowledge, our proposed algorithm, Policy Gradient Bayesian Robust Optimization for Imitation Learning (PG-BROIL), is the first policy optimization algorithm robust to a distribution of reward hypotheses that can scale to complex MDPs with continuous state and action spaces.

To evaluate PG-BROIL, we consider settings where there is uncertainty over the true reward function. We first examine the setting where we have an a priori distribution over reward functions and find that PG-BROIL is able to optimize policies that effectively trade-off between expected and worst-case performance. %Next we demonstrate that PG-BROIL combined with inverse reward design \cite{hadfield2017inverse} enables policy optimization that is robust to negative side-effects when seeking to optimize a hand-designed reward function.
Then, we leverage recent advances in efficient Bayesian reward inference~\cite{brown2020safe} to infer a posterior over reward functions from preferences over demonstrated trajectories. While other approaches which do not reason about reward uncertainty overfit to a single reward function hypothesis, PG-BROIL optimizes a policy that hedges against multiple reward function hypotheses. 
When there is high reward function ambiguity due to limited demonstrations, we find that PG-BROIL results in significant performance improvements over other state-of-the-art imitation learning methods.
% \ashwin{would be good to quantify performance improvement more concretely}

% demonstrate that PG-BROIL outperforms state-of-the-art imitation learning approaches when learning from preferences over demonstrations when the demonstrations lead to ambiguity about the true reward function. 

% We leverage recent advances in Bayesian reward inference \cite{brown2020safe} to efficiently infer a posterior over reward functions from preferences over offline demonstrations without requiring an MDP solver in the loop and then optimize a robust policy with respect to this posterior distribution. \adnote{explain better how?}

%%% While optimizing a policy to be robust to noise in dynamics is important, this type of uncertainty is aleatoric uncertainty---uncertainty that is inherent in the environment and cannot be reduced by learning more about the environment. While an agent can learn to act more cautiously or avoid areas with highly stochastic dynamics, this uncertainty never goes away. In this paper we are focused on dealing with epistemic uncertainty over the reward function---uncertainty that results from not having adequate information about the true reward function. Such uncertainty is reducible in principle, for example by collecting a much larger set of demonstrations which resolves any ambiguities in the demonstrators objective. However, this is often impractical, and it can be difficult for a human demonstrator to gauge whether their demonstrations are sufficient to prevent ambiguity in their objective.
\section{Related Work}
%We now briefly review related work in the areas of safe and robust reinforcement learning and imitation learning.
\paragraph{Reinforcement Learning:}
%Most RL algorithms simply try to maximize the expected return under a single reward function. 
There has been significant recent interest in safe and robust reinforcement learning~\cite{garcia2015comprehensive}; however, most approaches are only robust with respect to noise in transition dynamics and only consider optimizing a policy with respect to a single reward function.
%Need to do some more background reading for this section. Probably talk about how RL methods still have stochasticity in terms of aleatoric uncertainty and seek to minimize this whereas we deal with epistemic uncertainty.
Existing approaches reason about risk measures with respect to a single task rewards~\cite{risk-RL, risk-sensitive-RL, pg-conditional-values, WCPG}, establish convergence to safe regions of the MDP~\cite{SAVED, abc-lmpc}, or optimize a policy to avoid constraint violations~\cite{achiam2017constrained,safety-framework, recovery-rl}.

In this paper, we develop a reinforcement learning algorithm which reasons about risk with respect to a belief distribution over the task reward function. We focus on being robust to tail risk by optimizing for conditional value at risk~\cite{rockafellar2000optimization}. However, unlike prior work~\cite{risk-RL, risk-sensitive-RL, pg-conditional-values, tamar2015optimizing,WCPG, meanvarRL}, which focuses on risk with respect to a known reward function and stochastic transitions, we consider policy optimization when there is epistemic uncertainty over the reward function itself. We formulate a soft-robustness approach that blends optimizing for expected performance and optimizing for the conditional value at risk. Recent work also considers soft-robust objectives when there is uncertainty over the correct transition model of the MDP~\citep{lobo2020soft,russel2020entropic}, rather than uncertainty over the true reward function.

\paragraph{Imitation Learning:}
Imitation learning approaches vary widely in reasoning about reward uncertainty. Behavioral cloning approaches simply learn to imitate the actions of the demonstrator, resulting in quadratic regret~\cite{ross2010efficient}. DAgger~\cite{ross2011reduction} achieves sublinear regret by repeatedly soliciting human action labels in an online fashion. While there has been work on safe variants of DAgger~\cite{zhang2016query,hoque2021lazydagger}, these methods only enable robust policy learning by asymptotically converging to the policy of the demonstrator, and always assume access to an expert human supervisor. 

Inverse reinforcement learning (IRL) methods are another way of performing imitation learning~\cite{arora2018survey}, where the learning agent seeks to achieve better sample efficiency and generalization by learning a reward function which is then optimized to obtain a policy. However, most inverse reinforcement learning methods only result in a point-estimate of the demonstrator's reward function~\cite{abbeel2004apprenticeship, ziebart2008maximum,  fu2017learning,brown2019extrapolating}.
Risk-sensitive IRL methods \cite{lacotte2018risk, majumdar2017risk,Santara2018} assume risk-averse experts and focus on optimizing policies that match the risk-aversion of the demonstrator; however, these methods focus on the aleatoric risk induced by transition probabilities and there is no clear way to adapt risk-averse IRL to the Bayesian robust setting, where the objective is to be robust to epistemic risk over reward hypotheses rather than risk with respect to stochasticity in the dynamics. Bayesian IRL approaches explicitly learn a distribution over reward functions conditioned on the demonstrations, but usually only optimize a policy for the expected reward function or MAP reward function under this distribution \cite{ramachandran2007bayesian,choi2011map,brown2020safe}.
% \mdp{This seem to be a bit repetitive with the content that is in the introduction}

We seek to optimize a policy that is robust to epistemic uncertainty in the true reward function of an MDP. Prior work on robust imitation learning has primarily focused on maxmin approaches which seek to optimize a policy for an adversarial worst-case reward function \cite{syed2008apprenticeship,Ho2016,Regan2009,hadfield2017inverse,huang2018learning}. However, these approaches can learn overly pessimistic behaviors~\cite{brown2018efficient} and existing approaches assume discrete MDPs with known transition dynamics~\cite{syed2008apprenticeship,Regan2009,hadfield2017inverse} or require fully solving an MDP hundreds of times \cite{huang2018learning}, effectively limiting these approaches to discrete domains. Recently, \citep{brown2020bayesian} proposed a method for robust Bayesian optimization for imitation learning (BROIL), which optimizes a soft-robust objective that balances expected performance with conditional value at risk \cite{rockafellar2000optimization}. However, their approach is limited to discrete state and action spaces and known transition dynamics. By contrast, we derive a novel policy gradient approach which enables robust policy optimization with respect to reward function uncertainty for domains with continuous states and action and unknown dynamics.
%Adversarial imitation learning algorithms attempt to match the distribution of features or states reached by the learned policy and supervisor demonstrations by (1) learning reward functions which enforce the constraint that the demonstrations must attain higher reward than all other possible policies~\cite{abbeel2004apprenticeship, Syed2008, syed2008apprenticeship} or (2) learn policies such that the resulting trajectories cannot be distinguished from the demonstrations \cite{Ho2016}

% ... talk about GAIL and follow-on work and precursors. \cite{abbeel2004apprenticeship} \cite{Syed2008} \cite{syed2008apprenticeship} \cite{huang2018learning}, \cite{airl}.

%Inverse Reinforcement Learning methods seek to directly estimate the unknown reward function of the demonstrator. Many popular approaches ignore uncertainty over the reward function and simply learn a point estimate that maximizes the likelihood of the demonstrations, often while also maximizing the entropy of the resulting policy~\cite{ziebart2008maximum}. Other approaches take a Bayesian approach and seek to learn a distribution over reward functions conditioned on the demonstrations \cite{ramachandran2007bayesian}. Classical Bayesian IRL approaches are limited due to computational requirements, but recent work has shown that Bayesian reward inference can be scaled to solve complex visual imitation learning tasks such as learning to play Atari games from raw pixel demonstrations without access to the true reward function \cite{brown2020safe}.
\section{Preliminaries and Notation}
%Before describing our proposed method, we briefly introduce notation and review some of the concepts necessary to understand our approach.

\subsection{Markov Decision Processes}
We model the environment as a Markov Decision Process (MDP)~\cite{Puterman2005}. An MDP is a tuple $(\states, \actions, r, P, \gamma, p_0)$, with state space $\states$, action space $\actions$ , reward function $r: \states\times\actions \to \Real$, transition dynamics $P:\states \times \actions \times \states \to [0,1]$, discount factor $\gamma \in [0,1)$, and initial state distribution $p_0$.
We consider stochastic policies $\pi: \states \times \actions \to [0,1]$ which output a distribution over $\mathcal{A}$ conditioned on a state $s \in \mathcal{S}$. We denote the expected return of a policy $\pi$ under reward function $r$ as $\rho(\pi,r) = \Ex_{\tau \sim \pi_\theta}[r(\tau)]$. %When learning from demonstrations we denote the expert's policy by $\pi_E$.

\subsection{Distributions over Reward Functions}
We are interested in solving MDPs when there is epistemic uncertainty over the true reward function. When we refer to the reward function as a random variable we will use $R$, and will use $r$ to denote a specific model of the reward function. Reward functions are often parameterized as a linear combination of known features \cite{abbeel2004apprenticeship,ziebart2008maximum,sadigh2017active} or as a deep neural network \cite{Ho2016,fu2017learning}. Thus, we can model uncertainty in the reward function as a distribution over $R$, or, equivalently, as a distribution over the reward function parameters. This distribution could be a prior distribution $\Pr(R)$ that the agent learns from previous tasks \cite{xu2018learning}. Alternatively, the distribution could be the posterior distribution $\Pr(R\mid  D)$ learned via Bayesian inverse reinforcement learning \cite{ramachandran2007bayesian} given demonstrations $D$, the posterior distribution $\Pr(R \mid \mathcal{P}, D)$ given preferences $\mathcal{P}$ over demonstrations \cite{sadigh2017active,brown2020safe}, or the posterior distribution $\Pr(R\mid  r')$ learned via inverse reward design given a human-specified proxy reward $r'$
% \adnote{r', since it's a specific model not the random variable?}
\cite{hadfield2017inverse,ratner2018simplifying}. This distribution is typically only available via sampling techniques such as Markov chain Monte Carlo (MCMC) sampling \cite{ramachandran2007bayesian,hadfield2017inverse,brown2020safe}.

\subsection{Risk Measures}
\begin{figure}[t]
    \centering
    \includegraphics[width=\linewidth]{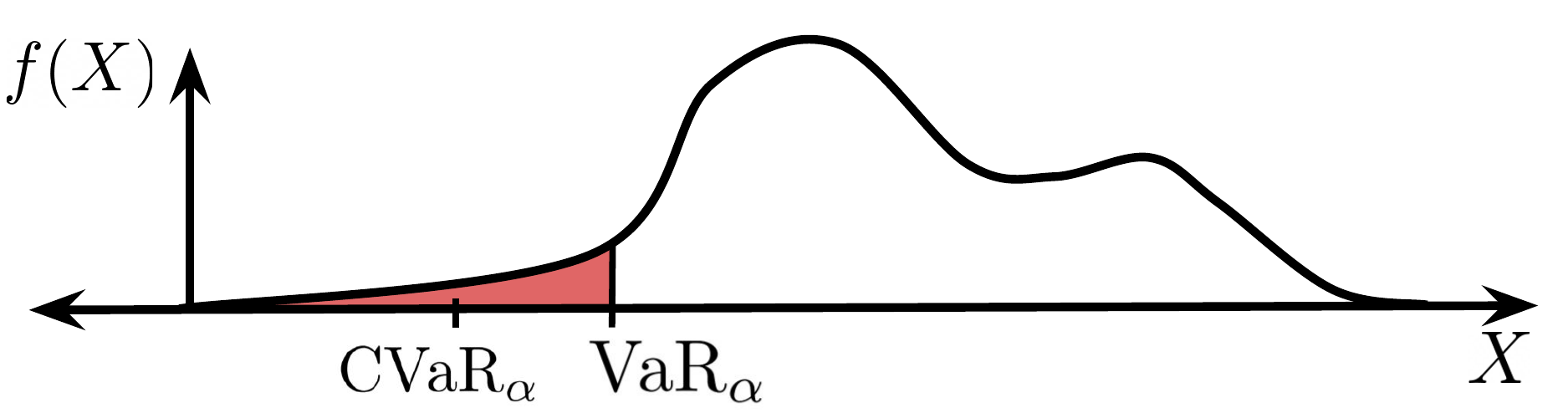}
    \caption{The pdf $f(X)$ of a random variable $X$. $\alphavar$ measures the $(1-\alpha)$-quantile outcome. $\alphacvar$ measures the expectation given that we only consider values less than the $\alphavar$.}
    \label{fig:cvar_var_example}
\end{figure}

%\DB{explain what alpha is}
We are interested in robust policy optimization with respect to a distribution over the performance of the policy induced by a distribution over possible reward functions. Consider a policy $\pi$ and a reward distribution $\Pr(R)$. Together, $\pi$ and $\Pr(R)$ induce a distribution over the expected return of the policy, $\rho(\pi, R), R\sim \Pr(R)$. We seek a robust policy that minimizes tail risk, given some risk measure, under the induced distribution $\rho$. Figure~\ref{fig:cvar_var_example} visualizes two common risk measures: value at risk ($\var$) and conditional value at risk ($\cvar$), for a general random variable $X$. In our setting, $X$ corresponds to the expected return, $\rho(\pi,R)$, of a policy $\pi$ under the reward function random variable $R$, and the objective is to minimize the tail risk (visualized in red).
% \adnote{We seek a robust policy that minimizes tail risk, under some measure of risk, under the induced distribution over $\rho$. Figure .. visulizes the risk measures we discuss next for a general random variable $X$ -- in our setting, $X$ corresponds to $\rho$.} Figure~\ref{fig:cvar_var_example} shows the distribution over a random variable $X$. In our case we are concerned with the distribution over the expected performance of our policy under $\Pr(R)$. 
% We seek a robust policy that minimizes tail risk (shown in red) under some measure of risk. In this section we briefly discuss two popular risk measures from finance: value at risk ($\var$) and conditional value at risk ($\cvar$).

\subsubsection{Value at Risk}
Given a risk-aversion parameter $\alpha \in [0,1]$, the $\alphavar$ of a random variable $X$ is the $(1-\alpha)$-quantile outcome:
\begin{equation}
\alphavar[X] = \sup \{x : \Pr(X \geq x) \geq \alpha\},
\end{equation}
where it is common to have $\alpha \in [0.9,1]$.
%\begin{wrapfigure}{r}{0.37\textwidth}

%\end{wrapfigure}
Despite the popularity of $\var$, optimizing a policy for $\var$ has several problems: (1) optimizing for $\var$ results in an NP hard optimization problem \cite{delage2010percentile}, (2) $\var$ ignores risk in the tail that occurs with probability less than $(1-\alpha)$ which is problematic for domains where there are rare but potentially catastrophic outcomes, and (3) $\var$ is not a coherent risk measure \cite{artzner1999coherent}. %In particular it violates the property that diversification leads to less risk.% (see the Appendix for a concrete example in terms of diversifying via a mixture policy).

\subsubsection{Conditional Value at Risk}
$\cvar$ is a coherent risk measure \cite{delbaen2002coherent}, also known as average value at risk, expected tail risk, or expected shortfall. For continuous distributions
\begin{equation}\label{eqn:cvar_cont}
\cvar_\alpha[X] = \Ex_{f(X)}\left[ X ~\mid ~ X \le \var_\alpha[X]\right].
\end{equation}
In addition to being coherent, $\cvar$ can be maximized via convex optimization, does not ignore the tail of the distribution, and is a lower bound on $\var$.
%Thus, $\cvar$ is often preferable over $\var$ \cite{rockafellar2000optimization}.
Because of these desirable properties, we would like to use $\cvar$ as our risk measure. 
%This allows us to learn policies which explicitly optimize for sufficiently high expected return under $\rho$ even for worst-case reward functions. 
%  when optimizing robust policies with respect to reward function uncertainty. \adnote{I think keep relating to $\rho$ so that peoople undestand what optimizing this is really saying -- maximizing this will mean over policies will mean what? give intuition}
 % is, whereas $\var$ only considers the quantile, but ignores any worse outcomes.
However, because posterior distributions obtained via Bayesian IRL are often discrete \cite{ramachandran2007bayesian,sadigh2017active,hadfield2017inverse,brown2018efficient}, we cannot directly optimize for CVaR using the definition in Equation~\eqref{eqn:cvar_cont} since this definition only works for atomless distributions. %\DB{Can we think of a one sentence intuition for why to help readers who aren't familiar with var and cvar?}
%However, when performing Bayesian inference over a reward function, we typically only have access to a finite number of samples from the posterior distribution over reward functions. %Thus, we are interested in maximizing $\cvar$ given finite number of samples.
Instead, we make use of the following definition of $\cvar$, proposed by ~\citet{rockafellar2000optimization}, that works for any distribution:
\begin{equation}\label{eq:cvar_convex}
     \alphacvar[X] = \max_{\sigma}\; \left( \sigma - \frac{1}{1-\alpha} \Ex [(\sigma  - X)_+] \right) ~,
\end{equation}
where $(x)_+ = \max(0,x)$ and $\sigma$ roughly corresponds to the $\alphavar$. To gain intuition for this formula, note that if we define $\sigma = \alphavar[X]$ we can rewrite $\alphacvar$ as
\begin{align}
    \alphacvar[X] &= \Ex_{f(X)}[X \mid X \leq \sigma] \\
    &= \sigma - \Ex_{f(X)}[\sigma - X \mid X \leq \sigma] \\
    &= \sigma - \frac{\Ex_{f(X)}[\mathbf{1}_{X \leq \sigma} \cdot (\sigma - X)]}{P(X \leq \sigma)} \\
    &= \sigma - \frac{1}{1 - \alpha}\Ex_{f(X)}[ (\sigma - X)_+]
\end{align}
where $\bm{1}_{x} = 1$ is the indicator function that evaluates to 1 if $x$ is True and 0 otherwise, and where we used the linearity of expectation, the definition of conditional expectation, and the definitions of $\alphavar[X]$, and $(x)_+$. Taking the maximum over $\sigma \in \Real$, gives us the definition in Equation~\eqref{eq:cvar_convex}.
\section{Bayesian Robust Optimization for Imitation Learning}
%\mdp{I am a little concerned that the contribution starts on page 5 of the paper. I am not sure if that is a big deal, but I have seen reviewers complain about this. One simple idea would be to move the related work to later in the paper.}
In Section~\ref{subsec:broil_objective} we describe the Bayesian robust optimization for imitation learning (BROIL) objective, previously proposed by \cite{brown2020bayesian}. Then, in sections~\ref{subsec:broil_pg} and~\ref{subsec:broil_pg_intuition}, we derive a novel policy gradient update for BROIL and provide an intuitive explanation for the result.

\subsection{Soft-Robust BROIL Objective}\label{subsec:broil_objective}
Rather than seeking a purely risk-sensitive or purely risk-neutral approach, we seek to optimize a soft-robust objective that balances the expected and probabilistic worst-case performance of a policy. Given some performance metric $\psi(\pi_\theta,R)$ where $R \sim \Pr(R)$, \citet{brown2020bayesian} recently proposed Bayesian Robust Optimization for Imitation Learning (BROIL) which seeks to optimize the following:
\begin{equation} \label{eq:broil_general}
\max_{\pi_\theta}  \lambda \cdot \Ex_{\Pr(R)}[\psi(\pi_\theta,R)] + (1-\lambda) \cdot \alphacvar \big[ \psi(\pi_\theta,R) \big]
\end{equation}
For MDPs with discrete states and actions and known dynamics, \citet{brown2020bayesian} showed that this problem can be formulated as a linear program which can be solved in polynomial time. However, many MDPs of interest involve continuous states and actions and unknown dynamics.

\subsection{BROIL Policy Gradient} \label{subsec:broil_pg}
We now derive a policy gradient objective for BROIL that allows us to extend BROIL to continuous states and actions and unknown transition dynamics, enabling robust policy learning in a wide variety of practical settings.
% We start with the same objective:
% \begin{align}
% \operatorname*{max}_{\pi_\theta} \quad &\lambda \Ex[\psi(\pi_\theta,R)] + (1-\lambda) \alphacvar \bigg[ \psi(\pi_\theta,R) \bigg]
% \end{align}
Given a parameterized policy $\pi_{\theta}$ and $N$ possible reward hypotheses, there are many possible choices for the performance metric $\psi(\pi_{\theta},R)$.~\citet{brown2020safe} considered two common metrics: (1) expected value, i.e., $\psi(\pi_{\theta},R) = \rho(\pi,R) = \Ex_{\tau \sim \pi_\theta}[R(\tau)]$ and (2) baseline regret, i.e., $\psi(\pi_{\theta},R) = \rho(\pi_\theta,R) - \rho(\pi_{E},R)$ where $\pi_E$ denotes an expert policy (usually estimated from demonstrations). In Appendix~\ref{app:broil_cvar_pg_proof} we derive a more general form for any performance metric $\psi(\pi_{\theta},R)$ and also give the derivation for the baseline regret performance metric.
For simplicity, we let $\psi(\pi_{\theta},R) = \rho(\pi,R)$ (expected return) hereafter.
% Thus, our objective is
% \begin{align}
% \max_{\pi_\theta} \quad &\lambda \Ex_{\Pr(R)}[\rho(\pi_\theta,R)] + (1-\lambda)  \alphacvar \bigg[ \rho(\pi_\theta,R)\bigg].
% \end{align}

To find the policy that maximizes Equation~\eqref{eq:broil_general} we need the gradient with respect to the policy parameters $\theta$. For the first term in Equation~\eqref{eq:broil_general}, we have
\begin{align}
    \nabla_\theta \Ex_{\Pr(R)}[\rho(\pi_\theta, R)]
    %&= \Ex_{\Pr(R)}[\nabla_\theta\Ex_{\tau \sim \pi_\theta}[R(\tau)]] \\
    &\approx \sum_{i=1}^{N} \Pr(r_i)\nabla_\theta\Ex_{\tau \sim \pi_\theta}[r_i(\tau)].
\end{align}

Next, we consider the gradient of the CVaR term. CVaR is not differentiable everywhere so we derive a sub-gradient. Given a finite number of samples from the reward function posterior, we can write this sub-gradient as
\begin{align}
\nabla_\theta  \max_\sigma \Bigl(\sigma -\frac{1}{1-\alpha} \sum_{i=1}^{N} \Pr(r_i)  \big(\sigma - \Ex_{\tau \sim \pi_\theta}[r_i(\tau)]  \big)_+ \Bigr)
\end{align}
where $(x)_+ = \max(0,x)$.
To solve for the sub-gradient of this term, note that given a fixed policy $\pi_\theta$, we can solve for $\sigma$ via a line search: since the objective is piece-wise linear we only need to check the value at each point $\rho(\pi,r_i)$, for each reward function sample from the posterior since these are the endpoints of each linear segment. If we let $\rho_i = \rho(\pi,r_i)$ then we can quickly iterate over all reward function hypotheses and solve for $\sigma$ as
\begin{equation} \label{eq:solve_for_sigma}
\sigma^* = \operatorname*{argmax}_{\sigma \in \{\rho_1,\ldots,\rho_N\}}  \Bigl(\sigma -\frac{1}{1-\alpha} \sum_{i=1}^{N} \Pr(r_i)  \big[\sigma - \rho_i  \big]_+ \Bigr).
\end{equation}
Solving for $\sigma^*$ requires estimating $\rho_i$ by collecting a set $\mathcal{T}$ of on-policy trajectories $\tau \sim \pi_\theta$ where $\tau = (s_0,a_0,s_1,a_1,\ldots,s_T,a_T)$:
\begin{equation} \label{eq:exp_return_hyps}
    \rho_i \approx\frac{1}{|\mathcal{T}|} \sum_{\tau \in \mathcal{T}} \sum_{t=0}^T r_i(s_t, a_t).
\end{equation}
Solving for $\sigma^*$ does not require additional data collection beyond what is required for standard policy gradient approaches. We simply evaluate the set of rollouts $\mathcal{T}$ from $\pi_{\theta}$ under each reward function hypothesis, $r_i$ and then solve the optimization problem above to find $\sigma^*$. While this requires more computation than a standard policy gradient approach---we have to evaluate each rollout under $N$ reward functions---this does not increase the online data collection, which is often the bottleneck in RL algorithms.%, adding more offline computation does not create a significant burden when compared with a standard risk-neutral policy gradient algorithm which evaluates rollouts under a single reward function. %Given $N$ sample reward functions this does require more reward evaluations and more computation than a standard policy gradient algorithm that only has one reward function, but these computations can be done without any access to the environment and only require function evaluations (perhaps by running the state-action pair through a deep neural network) so the added computational burden should not be significant when compared with the standard costs in RL of rolling out multiple trajectories in the actual MDP, especially for real world RL problems on an actual robot.

% \begin{equation}
%     \nabla_\sigma \left( \sigma -\frac{1}{1-\alpha} \sum_i \Pr(R_i) \big[\sigma - \Ex_{\tau \sim \pi_\theta}[R_i(\tau)]  \big]_+ \right) =
%     1 -\frac{1}{1-\alpha} \sum_i \Pr(R_i)  \bm{1}_{\sigma > \rho(\pi,R_i)}
% \end{equation}

% So it seems like we can solve for $\sigma$ by following this sub-gradient.

Given the solution $\sigma^*$ found by solving the optimization problem in~\eqref{eq:solve_for_sigma}, we perform a step of policy gradient optimization by following the sub-gradient of CVaR with respect to the policy parameters $\theta$:
% \mdp{what is $i$ in the sum below?}
\begin{align}
    \label{eq:subgrad}
    \nabla_\theta \alphacvar =
    \frac{1}{1-\alpha} \sum_{i=1}^{N} \Pr(r_i) \bm{1}_{\sigma^* \geq \rho(\pi_\theta,r_i)} \nabla_\theta\rho(\pi_\theta, r_i)
\end{align}
where $\bm{1}_x$ is the indicator function that evaluates to 1 if $x$ is True and 0 otherwise.
%     0 & \text{otherwise}
% \begin{equation}
%     \bm{1}_x =
%     \begin{cases}
%     1 &\text{if $x$ is True}\\
%     0 & \text{otherwise}
%     \end{cases}
% \end{equation}
Given the sub-gradient of the BROIL objective~\eqref{eq:subgrad}, the only thing remaining to compute is the standard policy gradient.
Note that in standard RL, we write the policy gradient as~\cite{sutton2018reinforcement}:
\begin{align}
    \nabla_\theta \Ex_{\tau \sim \pi_\theta} [ R(\tau)] = \Ex_{\tau \sim \pi_\theta} \left[ \sum_{t=0}^T \nabla_\theta \log \pi_\theta(a_t \mid s_t) \Phi_t(\tau) \right]
\end{align}
where $\Phi_t$ is a measure of the performance of trajectory $\tau$ starting at time $t$. One of the most common forms of $\Phi_t(\tau)$ is the on-policy advantage function \citep{schulman2015high} with respect to some single reward function:
\begin{equation}
    \Phi_t(\tau) = A^{\pi_\theta}(s_t, a_t) = Q^{\pi_\theta}(s_t, a_t) - V^{\pi_\theta}(s_t).
\end{equation}
If we define $\Phi^{r_i}_t$ in terms of a particular reward function $r_i$, then, as we show in Appendix~\ref{app:broil_cvar_pg_proof}, we can rearrange terms in the standard policy gradient formula to obtain the following form for the BROIL policy gradient which we estimate using a set $\mathcal{T}$ of on-policy trajectories $\tau \sim \pi_\theta$ where $\tau = (s_0,a_0,s_1,a_1,\ldots,s_T,a_T)$ as follows:
\begin{align}\label{eq:pg_broil_general}
\nabla_\theta \text{BROIL}
\approx& \frac{1}{|\mathcal{T}|} \sum_{\tau \in \mathcal{T}}  \biggl[ \sum_{t=0}^T \nabla_\theta \log \pi_\theta(a_t \mid s_t) w_t(\tau) \biggr]
\end{align}
where
\begin{equation}\label{eq:broil_pg_weight}
w_t(\tau) = \sum_{i=1}^{N}\Pr(r_i)\Phi_t^{r_i}(\tau)\left(\lambda   +  \frac{1-\lambda}{1-\alpha} \bm{1}_{\sigma^* \geq \rho(\pi,r_i)} \right)
\end{equation}
is the weight associated with each state-action pair $(s_t, a_t)$ in the set of trajectory rollouts $\mathcal{T}$. The resulting vanilla policy gradient algorithm is summarized in Algorithm~\ref{alg:pgbroil}. In Appendix~\ref{app:ppo_broil} we show how to apply a trust-region update based on Proximal Policy Optimization \cite{schulman2017proximal} for more stable policy gradient optimization.

\subsection{Intuitive Interpretation of the Policy Gradient}\label{subsec:broil_pg_intuition}
Consider the policy gradient weight $w_t$ given in Equation~\eqref{eq:broil_pg_weight}.
If $\lambda=1$, then
\begin{equation}
w_t(\tau) = \sum_{i=1}^{N} \Pr(R_i)\Phi_t^{R_i}(\tau) = \Phi_t^{\bar{R}}(\tau)
\end{equation}
where $\bar{R}$ is the expected reward under the posterior. Thus, $\lambda=1$ is equivalent to standard policy gradient optimization under the mean reward function and gradient ascent will focus on increasing the likelihood of actions that look good in expectation over the reward function distribution $\Pr(R)$. Alternatively, if $\lambda=0$, then
\begin{equation}
w_t(\tau) = \frac{1}{1-\alpha} \sum_{i=1}^{N}\bm{1}_{\sigma^* \geq \rho(\pi,R_i)} \Pr(R_i) \Phi_t^{R_i}(\tau)
\end{equation}
and gradient ascent will increase the likelihood of actions that look good under reward functions that the current policy $\pi_\theta$ performs poorly under, i.e., policy gradient updates will focus on improving performance under all $R_i$ such that $\rho(\pi,R_i) \leq \sigma^*$, weighting the gradient according to the likelihood of these worst-case reward functions. The update rule also multiplies by $1/(1-\alpha)$ which acts to normalize the magnitude of the gradient: as $\alpha \rightarrow 1$ we update on reward functions further into the tail, which have smaller probability mass. Thus, $\lambda \in [0,1]$ allows us to blend between maximizing policy performance in expectation versus worst-case and $\alpha \in [0,1)$ determines how far into the tail of the distribution to focus the worst-case updates.

\begin{algorithm}[tb]
\caption{Policy Gradient BROIL}
\label{alg:pgbroil}
\begin{algorithmic}[1]
 \STATE {\bfseries Input:} initial policy parameters $\theta_0$, samples from reward function posterior $r_1,\ldots,r_N$ and associated probabilities, $\Pr(r_1),\ldots,\Pr(r_N)$.

    \FOR{$k=0,1,2,\ldots$}
        \STATE Collect set of trajectories $\mathcal{T}_k = \{ \tau_i \}$ by running policy $\pi_{\theta_k}$ in the environment.
        \STATE Estimate expected return of $\pi_{\theta_k}$ under each reward function hypothesis $r_j$ using~Eq.~\eqref{eq:exp_return_hyps}.
%             \begin{equation*}
%     \hat{\rho}_i = \frac{1}{|\mathcal{T}|} \sum_{\tau \in \mathcal{T}} \sum_{t=0}^T R_i(s_t, a_t).
% \end{equation*}
        \STATE Solve for $\sigma^*$ using Eq.~\eqref{eq:solve_for_sigma}
        % \begin{equation*}
        % \sigma^* = \operatorname*{argmax}_{\sigma \in \{\hat{\rho}_1, \ldots, \hat{\rho}_N \}}  \Bigl(\sigma -\frac{1}{1-\alpha} \sum_i \Pr(R_i)  \big[\sigma - \hat{\rho}_i  \big]_+ \Bigr)
        % \end{equation*}
    \STATE Estimate policy gradient using Eq.~\eqref{eq:pg_broil_general} and Eq.~\eqref{eq:broil_pg_weight}.
        % \begin{equation*}
        %     \hat{g}_k = \frac{1}{|\mathcal{T}|} \sum_{\tau \in \mathcal{T}} \bigg[ \sum_{t=0}^T \nabla_\theta \log \pi_{\theta_k}(a_t \mid s_t) w_t \bigg]
        % \end{equation*}
        % where $w_t = \sum_i \Pr(R_i)\Phi_t^{R_i}(\tau)\big(\lambda   +  \frac{1-\lambda}{1-\alpha} \bm{1}_{\sigma^* > \hat{\rho}_i)} \big)$.
        \STATE Update $\theta$ using gradient ascent.
        % \begin{equation*}
        %     \theta_{k+1} = \theta_k + \alpha_k \hat{g}_k.
        % \end{equation*}
        % for some learning rate $\alpha_k$.% (e.g. via Adam \citep{kingma2014adam}).
    \ENDFOR
\end{algorithmic}
\end{algorithm}

% \vspace{-.1cm}
\section{Experiments}
In experiments, we consider the following questions: (1) Can PG-BROIL learn control policies in MDPs with continuous states and actions and unknown transition dynamics? (2) Does optimizing PG-BROIL with different values of $\lambda$ effectively trade-off between maximizing for expected return and maximizing robustness? (3) When demonstrations are ambiguous, can PG-BROIL outperform other imitation learning baselines by hedging against uncertainty? 

Code and videos are available at \url{https://sites.google.com/view/pg-broil}.

\subsection{Prior over Reward Functions}\label{sec:prior_dist}
We first consider an RL agent with a priori uncertainty over the true reward function. This setting allows us to initially avoid the difficulties of inferring a posterior distribution over reward functions and carefully examine whether PG-BROIL can trade-off expected performance and robustness (CVaR) under epistemic uncertainty over the true reward function. We study 3 domains: the classical CartPole benchmark~\cite{gym}, a pointmass navigation task inspired by~\cite{SAVED} and a robotic reaching task from the from the DM Control Suite~\cite{tassa2020dmcontrol}. All domains are characterized by a robot navigating in an environment where some states have uncertain costs. All domains have unknown transition dynamics and continuous states and actions (except CartPole which has discrete actions). We implement PG- BROIL on top of OpenAI Spinning Up \cite{spinning-up}. For cartpole we implement PG-BROIL on top of REINFORCE \cite{vanilla-pg} and for remaining domains we implement PG-BROIL on top of PPO~\cite{schulman2017proximal} (see Appendix~\ref{app:ppo_broil}). 
% For all experiments, we implement PG-BROIL on top of the S implementation of the REINFORCE~\cite{vanilla-pg} (CartPole environment) and PPO~\cite{schulman2017proximal} (Pointmass Navigation and Reacher environments) implementations provided in~\cite{spinning-up}.

\begin{figure*}
    \centering
    \includegraphics[width=\linewidth]{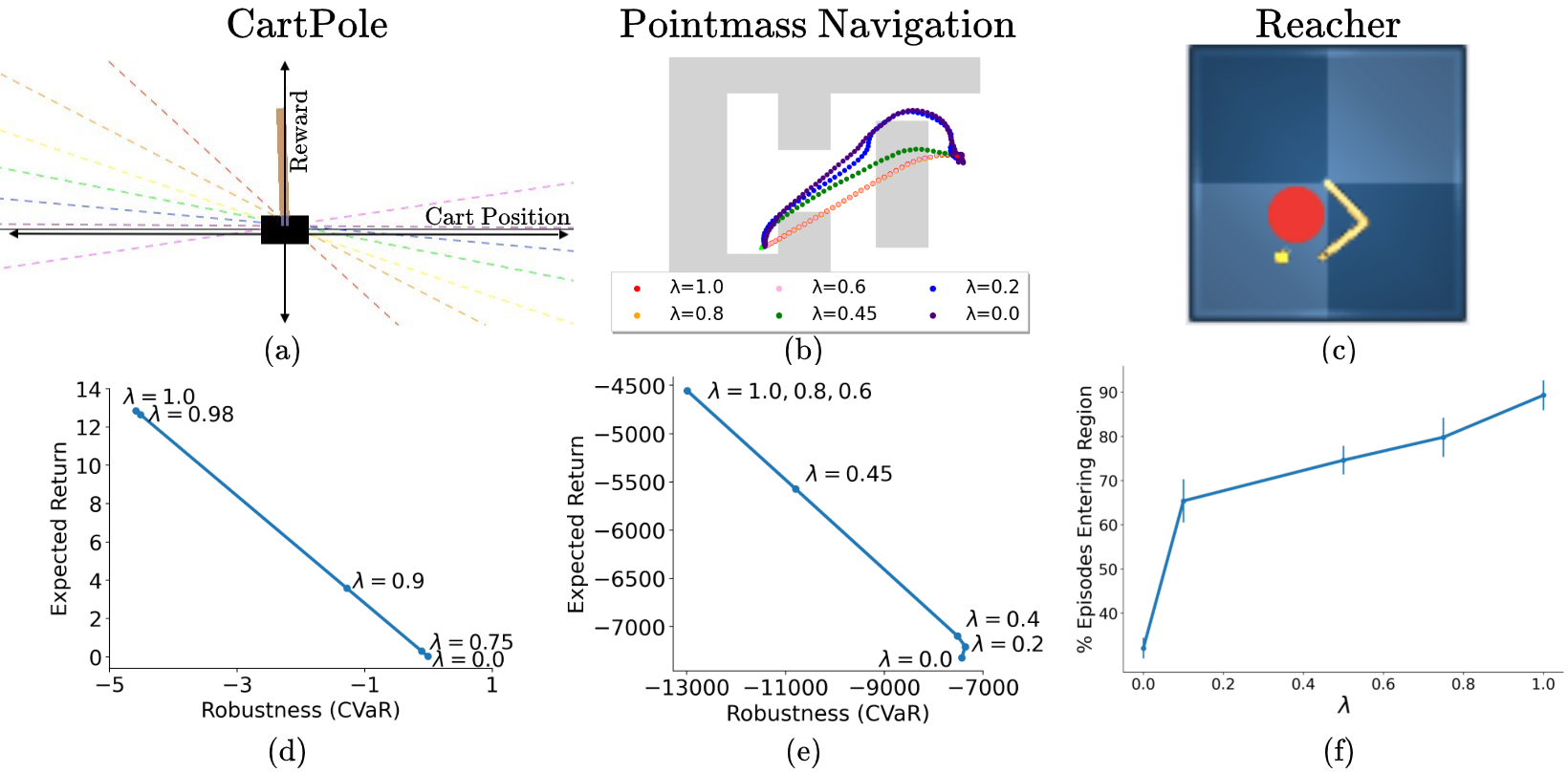}
    \caption{\textbf{Prior over Reward Functions: Domains and Results.} We study (a) CartPole in which the reward is an unknown linear function of the cart's position, (b) Pointmass Navigation with gray regions of uncertain costs, and (c) Reacher with a red region of uncertain cost. For the CartPole and Pointmass Navigation domains, we find that as $\lambda$ is decreased, the learned policy optimizes more for being robust to tail risk and thus achieves more robust performance (in terms of CVaR) at the expense of expected return in panels (d) and (e). In panel (f), we find that the reacher arm enters the riskier red region less often with decreasing $\lambda$ as expected.}
    \vspace{-0.1in}
    \label{fig:reward-priors-domains}
\end{figure*}

\subsubsection{Experimental Domains}
\textbf{CartPole: }We consider a risk-sensitive version of the classic CartPole benchmark \cite{gym}. The reward function is $R(s) = b \cdot s_x$, where $s_x$ is the position of the cart on the track, and there is uncertainty over $b$. Our prior over $b$ is distributed uniformly in the range [-1, 0.2]. The center of the track is $s_x=0$. We sample values of $b$ between -1 and 0.2 across even intervals of 0.2 width to form a discrete posterior distribution for PG-BROIL. The reward distribution is visualized in Figure~\ref{fig:reward-priors-domains}a. Based on our prior distribution over reward functions, the left side of the track ($s_x < 0$) is associated with a higher expected reward but a worse worst case scenario (the potential for negative rewards). By contrast, the robust solution is to stay in the middle of the track in order to perform well across all possible reward functions since the center of the track has less risk of a significantly negative reward than the left or right sides of the track. 
%While simple, this simulates more complicated settings where behavior that maximizes only expected reward can lead to more variance, whereas a risk-sensitive policy makes up for losses in expectation by having much smaller tails with respect to the expected performance over the reward function posterior. 
 %We see that most of the reward functions are maximized on the left side of the track, however 
 
\textbf{Pointmass Navigation: }We next consider a risk-sensitive continuous 2-D navigation task inspired by~\citet{SAVED}. Here the objective is to control a pointmass robot towards a known goal location with forces in cardinal directions in a system with linear Gaussian dynamics and drag. There are gray regions of uncertain cost that can either be traversed or avoided as illustrated in Figure~\ref{fig:reward-priors-domains}b. For example, these regions could represent grassy areas which are likely easy to navigate, but where the grass may occlude mud or holes which would impede progress and potentially cause damage or undue wear and tear on the robot. The robot has prior knowledge that it needs to reach the goal location $g = (0,0)$ on the map, depicted by the red star. We represent this prior with a nominal cost for each step that is the distance to the goal from the robot's position. We add a penalty term of uncertain cost for going through the gray region giving the following reward function posterior:
% \ashwin{this reward function is written as a cost, it should probably be all negated right?}
\begin{equation}
    R(s) = -\left(\|s_{x,y} - g \|_2^2 + b \cdot \mathbf{1}_{\rm gray}\right), b \sim \Pr(b),
\end{equation}
where $\mathbf{1}_{\rm gray}$ is an indicator for entering a gray region, and where the distribution $\Pr(b)$ over the penalty $b$ is given as
\begin{center}
\begin{tabular}{|c|ccccc|}
\hline
  $b$   & -500 & -40 & 0 & 40 & 50 \\
  \hline
  $\Pr(b)$ & 0.05 & 0.05 & 0.2 & 0.3 & 0.4 \\
  \hline
\end{tabular}
\end{center}
On average it is favorable to go through the gray region ($\Ex[b] = +5$), but there is some probability that going through the gray region is highly unfavorable:

\textbf{Reacher: }We design a modified version of the Reacher environment from the DeepMind Control Suite \cite{tassa2020dmcontrol} (Figure~\ref{fig:reward-priors-domains}c), which is a 2 link planar arm where the robot can apply joint torques to each of the 2 joints to guide the end effector of the arm to a goal position on the plane. We modify the original environment by including an area of uncertainty (large red circle). When outside the uncertain region, the robot receives a reward which penalizes the distance between the end effector and the goal (small yellow circle). Thus, the robot is normally incentivized to guide the end effector to the goal as quickly as possible. When the end effector is inside the uncertain region, the robot has an 80\% chance of receiving a +2 bonus, a 10\% chance of receiving a -2 penalty, and a 10\% chance of neither happening (receiving rewards as if it were outside the uncertain region). The large red circle can be interpreted as a region on the table that has a small chance of causing harm to the robot or breaking an object on the table. However, in expectation the robot believes it is good to enter the red region (e.g., assuming that objects in this region are not fragile).

\subsubsection{Results}
PG-BROIL consistently exhibits more risk-averse behaviors with decreasing $\lambda$ across all domains.
For CartPole and Pointmass Navigation, we see that as $\lambda$ is decreased, the learned policy becomes more robust to tail risk at the expense of lower expected return in Figures~\ref{fig:reward-priors-domains}d and~\ref{fig:reward-priors-domains}e respectively. Figure~\ref{fig:reward-priors-domains}e indicates that values of $\lambda$ close to 0 can lead to unstable policy optimization due to excessive focus on tail risk---the policy for $\lambda=0$ is Pareto dominated by the policy for $\lambda=0.2$. We visualize the learned behaviors for different values of $\lambda$ for the Pointmass Navigation environment in Figure~\ref{fig:reward-priors-domains}b. For high values of $\lambda$, the robot cuts straight through the uncertain terrain, for intermediate values (eg. $\lambda = 0.45$), the robot somewhat avoids the uncertain terrain, while for low values of $\lambda$, the robot almost entirely avoids the uncertain terrain at the expense of a longer path. 
Finally, for the Reacher environment, we find that the percentage of episodes where the arm enters the red region decreases as $\lambda$ decreases as expected (Figure~\ref{fig:reward-priors-domains}f).

\subsection{Learning from Demonstrations}\label{sec:brex}
Our previous results demonstrated that PG-BROIL is able to learn policies that effectively balance expected performance and robustness in continuous MDPs under a given prior over reward functions. In this section, we consider the imitation learning setting where a robot infers a reward function from demonstrated examples. Given such input, there are typically many reward functions that are consistent with it; however, many reward inference algorithms~\cite{fu2017learning, finn2016guided, brown2019extrapolating} will output only one of them---not necessarily the true reward. There has been some work on Bayesian algorithms such as Bayesian IRL~\cite{ramachandran2007bayesian} which estimates a \emph{posterior distribution} instead of a single reward and Bayesian REX~\cite{brown2020safe} which makes it possible to efficiently learn this posterior from preferences over high dimensional demonstrated examples of varying qualities. However, prior work on Bayesian reward learning often only optimizes policies for the expected or MAP reward estimate over the learned posterior~\cite{ramachandran2007bayesian,choi2011map,brown2020safe}. Our hypothesis is that for imitation learning problems with high uncertainty about the true reward function, taking a robust optimization approach via PG-BROIL will lead to better performance by producing policies that do well in expectation, but also avoid low reward under \emph{any} of the sufficiently probable reward functions in the learned posterior.

\begin{figure*}[th!]
    \centering
    \includegraphics[width=\linewidth]{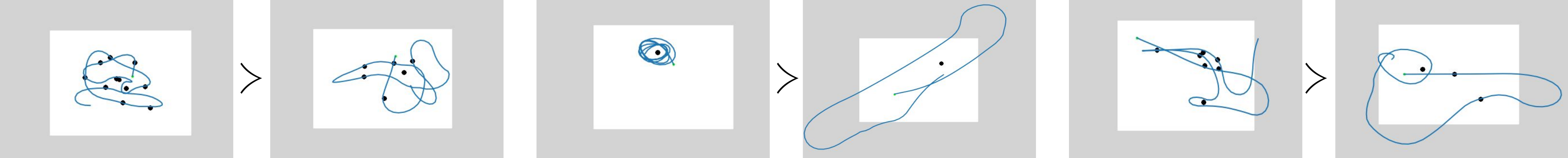}
    \caption{\textbf{TrashBot environment: } %The robot is tasked with picking up pieces of trash (black dots) while avoiding the gray regions. We illustrate an example trajectory from a learned policy in blue. 
    Each time the robot picks up a piece of trash (by moving close to a black dot), a new one appears at a randomly in the white region. We give pairwise preferences over human demos that aim to teach the robot that picking up trash is good (left), going into the gray region is undesirable (center), and less time in the gray region and picking up more trash is preferred (right). 
    }
    \label{fig:trashbot}
    \vspace{-0.1in}
\end{figure*}

\subsubsection{TrashBot from Demos}
We first consider a continuous control TrashBot domain (Figure~\ref{fig:trashbot}), where aim to teach a robot to  pick up pieces of trash (black dots) while avoiding the gray boundary regions. The state-space, dynamics and actions are the same as for the Pointmass Navigation environment and we provide human demonstrations via a simple teleoperation interface. The robot constructs its reward function hypotheses as linear combinations of three binary features which correspond to: (1) being in the gray region (GRAY), (2) being in the white region (WHITE), and (3) picking up a piece of trash (TRASH). We give three pairwise preferences over human teleoperated trajectories (generated by one of the authors) as shown in Figure~\ref{fig:trashbot}. However, the small number of preferences makes it challenging for the robot to ascertain the true reward function parameters as there are many reward function weights that would lead to the same human preferences. Furthermore, the most salient feature is WHITE and this feature is highly correlated, but not causal, with the preferences. Thus, this domain can easily lead to reward hacking/gaming behaviors~\cite{reward-hacking}. We hypothesize that PG-BROIL will hedge against uncertainty and learn to pick up trash while avoiding the gray region.
% \vspace{-0.15in}
\begin{table}[t]
\caption{\textbf{TrashBot:} We evaluate PG-BROIL against 5 other imitation learning algorithms when learning from ambiguous preferences over demonstrations (Figure~\ref{fig:trashbot}). Results are averages ($\pm$ one st. dev.) over 10 random seeds and 100 test episodes each with a horizon of 100 steps per episode. For PG-BROIL, we set $\alpha=0.95$ and report results for the best $\lambda$ ($\lambda=0.8$). }
\label{sample-table}
\vskip 0.15in
\begin{center}
\begin{small}
\begin{sc}
\begin{tabular}{lrr}
\toprule
algorithm & \begin{tabular}[r]{@{}r@{}}Avg. Trash\\ Collected\end{tabular} & \begin{tabular}[r]{@{}r@{}}Avg. Steps in \\ Gray Region\end{tabular} \\
\midrule
BC & 3.4 $\pm$ 1.8 & 2.7  $\pm$ 6.2 \\
GAIL & 2.2  $\pm$ 1.5 & 3.7  $\pm$ 9.9 \\
RAIL & 1.1 $\pm$ 1.2 & 2.2 $\pm$ 6.9 \\
PBRL & 2.6  $\pm$ 1.5 & 1.2   $\pm$ 2.7 \\
Bayesian REX & 1.6 $\pm$ 1.3 & 1.2 $\pm$ 1.7 \\
\textbf{PG-BROIL}  & \textbf{8.4  $\pm$ 0.5} & \textbf{0.1  $\pm$ 0.1} \\
\bottomrule
\end{tabular}
\end{sc}
\end{small}
\end{center}
\vskip -0.1in
\end{table}

We compare against behavioral cloning (BC), GAIL~\cite{Ho2016}, and Risk-Averse Imitation Learning (RAIL)~\cite{Santara2018}, which estimates CVaR over trajectories to create a risk-averse version of the GAIL algorithm. To facilitate a fairer comparison, we only give BC, GAIL, and RAIL the better ranked demonstration from each preference pair. We also compare with Preference-based RL (PBRL)~\cite{christiano2017deep} in the offline demonstration setting \cite{brown2019extrapolating} which optimizes an MLE estimate of the reward weights
% , which optimizes the MLE estimate of the reward weights given the preferences: 
% %\begin{equation}
%     $\wb_{\rm MLE} = (-0.09, 0.09, 0.99)$,
%\end{equation}
and Bayesian REX~\cite{brown2020safe}, which optimizes the mean reward function under the posterior distribution given the preferences. PG-BROIL also uses Bayesian REX \citep{brown2020safe} to infer a reward function posterior distribution given the preferences over demonstrations (see Appendix~\ref{app:brex_details} for details), but optimizes the BROIL objective.

Table \ref{sample-table} compares the performance of each baseline imitation learning algorithm when given the 3 pairs of demonstrations shown in Figure~\ref{fig:trashbot}. %, on a robot trash picking environment.
We find that PG-BROIL outperforms BC and  GAIL~\cite{Ho2016} by not directly seeking to imitate the states and actions in the demonstrations, but by explicitly reasoning about uncertainty in the true reward function. We also find that PG-BROIL significantly outperforms RAIL. This is because RAIL only focuses on minimizing aleatoric uncertainty under stochastic transition dynamics for a single reward function (the discriminator), not epistemic uncertainty over the true reward function. We find that PG-BROIL outperforms PBRL and Bayesian REX.
%  however, these methods still suffer from only trying to mimic the demonstrations and overfit to the suboptimalities (see Appendix~\ref{app:trashbot_details}). 
%  and 
% :
% %\begin{equation}
%     $\wb_{\rm mean} = (-0.36, 0.34, 0.45)$,
%\end{equation}
%where the weights correspond to WHITE, GRAY, and TRASH features, respectively. 
% \db{TODO: figure out an intuitive explanation for this...}

We inspected the learned reward functions and found that 
the PBRL reward places heavy emphasis on collecting trash but has a small positive weight on the WHITE feature. We hypothesize that this results in policy optimization falling into a local maxima in which it mostly mines rewards by staying in the white region. By contrast, PG-BROIL considers a number of reward hypotheses, many of which have negative weights on the WHITE feature. Thus, a risk-averse agent cannot mine rewards by simply staying in the white region, and is incentivized to maximally pick up trash while keeping visits to the gray region low. %See \ashwin{insert ref to appendix} for further details.
% \ashwin{refer to hist over white feature weights learned by BREX to substantiate this claim}
The mean reward function optimized by Bayesian REX penalizes visiting the gray region but learns roughly equal weights for the WHITE and TRASH features. Thus, Bayesian REX is not strongly incentivized to pick up trash. Because of this the learned policy sometimes visits the borders of the white region and occasionally enters the gray region when it accumulates too high of a velocity. By contrast, PG-BROIL effectively optimizes a policy that is robust to multiple hypotheses that explain the rankings: picking up trash more than any other policy, while avoiding the gray region. 
See Appendix~\ref{app:trashbot_analysis}.

\subsubsection{Reacher from Demos with Domain Shift}
For this experiment, we use the same Reacher environment described above. We give the agent five pairwise preferences over demonstrations of varying quality in a training domain where the uncertain reward region is never close to the goal and where none of the demonstrations show the reacher arm entering the uncertain region. We then introduce domain shift by both optimizing and testing policies in reacher environments unseen in the demonstrations, where the goal location is randomized and sometimes the uncertain reward region is in between the the reacher arm and the goal.
% and use Bayesian REX \cite{brown2020bayesian} to infer posterior over reward functions for PG-BROIL. 
% All demonstrations reach the target and stay there for various amounts of time, except for one which never reaches the target. 
The inferred reward function is a linear combination of 2 features: TARGET and UNCERTAIN REGION which are simply binary indicators which identify whether the agent is in the target location or in the uncertain region respectively. In the posterior generated using Bayesian REX, we find that the weight learned for the TARGET feature is strongly positive over all reward functions. UNCERTAIN REGION, having no information from any of the demonstrations, has a wide variety of possible values from -1 to +1 (reward weights are normalized to have unit L2-norm). Both the mean and MLE reward functions assign a positive weight to both the TARGET and UNCERTAIN REGION features, resulting in Bayesian REX and PBRL frequently entering the uncertain region as shown in Table~\ref{reacher_demos_table}. By contrast, PG-BROIL hedges against its uncertainty over the quality of the uncertain region and avoids it.
% Similar to what we see in the TrashBot experiment in Section \ref{sec:brex}, the PBRL agent mines the positive reward from UNCERTAIN REGION, staying in the uncertain region more than attempting to reach the target. The Bayesian REX agent visits the target but also places the end effector within the uncertain region since it does not reason about the risk associated with entering the uncertain region. 
% \zaynah{improve explanation}
% For this experiment, we use the same hyperparameters and PPO implementation described in Appendix \ref{app:reacher_details}, but instead train the PPO agent for 250 epochs.
See Appendix~\ref{app:reacher_details}.

\begin{table}
\caption{\textbf{Reacher from Demos:} We evaluate PG-BROIL and baseline imitation learning algorithms when learning from preferences over demonstrations. Results are averages ($\pm$ one st. dev.) over 3 seeds and 100 test episodes with a horizon of 200 steps per episode. For PG-BROIL, we set $\alpha=0.9$ and report results for $\lambda=0.15$.}
\label{reacher_demos_table}
\vskip 0.15in
\begin{center}
\begin{small}
\begin{sc}
\begin{tabular}{lrr}
\toprule
algorithm & \begin{tabular}[r]{@{}r@{}}Avg. Steps in \\ Uncertain Region\end{tabular} & \begin{tabular}[r]{@{}r@{}}Avg. Steps in \\ Target Region\end{tabular} \\
\midrule
BC & 11.3  $\pm$ 27.4 & 39.9 $\pm$ 62.3  \\
GAIL & 2.3  $\pm$ 1.7 & 5.1  $\pm$ 13.0 \\
RAIL & 2.1 $\pm$ 1.2 & 4.6 $\pm$ 27.0 \\
PBRL & 28.4  $\pm$ 37.7 & 16.8   $\pm$ 30.4 \\
Bayesian REX & 13.5 $\pm$ 35.0 & 94.5  $\pm$ 70.1 \\
\textbf{PG-BROIL}  & \textbf{1.7  $\pm$ 7.2} & \textbf{102.0  $\pm$ 60.5} \\
\bottomrule
\end{tabular}
\end{sc}
\end{small}
\end{center}
\vskip -0.2in
\end{table}

\subsubsection{Atari Boxing from Demos}
\begin{figure}
     \centering
     \begin{subfigure}[b]{0.3\linewidth}
         \centering
         \includegraphics[width=\linewidth]{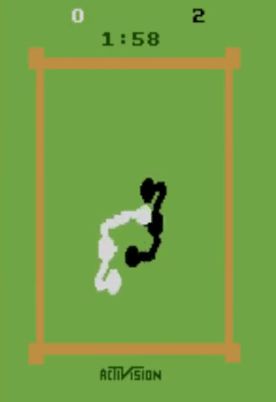}
         \caption{}
         \label{fig:y equals x}
     \end{subfigure}
     \hfill
     \begin{subfigure}[b]{0.69\linewidth}
        \begin{center}
        \begin{small}
        \begin{sc}
        \begin{tabular}{lrr}
        \toprule
        algorithm & \begin{tabular}[c]{@{}r@{}}Game Score\end{tabular} &  \\
        \midrule
        BC & 1.7  $\pm$ 5.3 &  \\
        GAIL & -0.2  $\pm$ 5.8 &  \\
        RAIL & 0.5 $\pm$ 4.9 &  \\
        PBRL & -15.0  $\pm$ 8.2 &  \\
        Bayesian REX & 1.6 $\pm$ 4.7 \\
        \textbf{PG-BROIL}  & \textbf{23.9  $\pm$ 13.5} & \\
        \bottomrule
        \end{tabular}
        \end{sc}
        \end{small}
        \end{center}
        \caption{}
     \end{subfigure}
        \caption{\textbf{Atari Boxing: } We evaluate PG-BROIL against baseline imitation learning algorithms when learning from preferences over demonstrations. Results are averages ($\pm$ one st. dev.) over 3 random seeds and 100 test episodes. For PG-BROIL, we set $\alpha=0.9$ and report results for the best $\lambda$ ($\lambda=0.3$). The game score is the number of hits the trained agent (white) scored minus the number of times the agent gets hit by the opponent (black).}
        \label{boxing_demos_table}
\end{figure}

For this experiment, we give the agent 3 preferences over suboptimal demos of the Atari Boxing game~\cite{bellemare2013arcade}. We use Bayesian REX to infer a reward function posterior where each inferred reward functions is a linear combinations of 3 binary indicator features identifying whether the agent hit its opponent, got hit, or stayed away from the opponent. The mean and MLE reward functions both assign a high weight to hitting the opponent, ignoring the risk of getting hit by the opponent due to always staying close to the opponent in order to score hits on it. PG-BROIL tries to satisfy multiple reward functions by both trying to avoid getting hit and scoring hits, resulting in better performance under the true reward as shown in Table~\ref{boxing_demos_table}. 
% We use the hyperparameters and PPO implementation described in \ref{app:boxing_details} and train for 800 epochs.
See Appendix~\ref{app:boxing_details} for more details.

% \vspace{-0.3in}

% \begin{table}
% \caption{\textbf{Atari Boxing Learning from Demonstrations: } We evaluate PG-BROIL against 3 other imitation learning algorithms when learning from ambiguous demonstrations. Results are averages ($\pm$ one st. dev.) over 3 random seeds and 100 test episodes. For PG-BROIL, we set $\alpha=0.9$ and report results for the best $\lambda$ ($\lambda=0.3$). The game score is the number of hits the trained agent (white) scored minus the number of times the agent gets hit by the opponent (black).}
% \includegraphics[scale=0.35]{figs/boxing_thumbnail.jpg}
% \includegraphics[scale=0.8]{figs/boxing_table.png}
% \label{boxing_demos_table}
% \vskip 0.15in
% %\begin{center}
% %\begin{small}
% %\begin{sc}
% %\begin{tabular}{lll}
% %\toprule
% %algorithm & \begin{tabular}[c]{@{}l@{}}Game Score\end{tabular} &  \\
% %\midrule
% %BC & x  $\pm$ x &  \\
% %GAIL & x  $\pm$ x &  \\
% %PBRL & -15.0  $\pm$ 8.15 &  \\
% %Bayesian REX & 1.62 $\pm$ 4.71 \\
% %RAIL & x $\pm$ x &  \\
% %\textbf{PG-BROIL}  & \textbf{23.9  $\pm$ 13.5} & \\
% %\bottomrule
% %\end{tabular}
% %\end{sc}
% %\end{small}
% %\end{center}
% \vskip -0.1in
% \end{table}
\section{Discussion and Future Work}

\textbf{Summary:} We derive a novel algorithm, PG-BROIL, for safe policy optimization in continuous MDPs that is robust to epistemic uncertainty over the true reward function. 
% The key insight is that the conditional value at risk (CVaR) \cite{rockafellar2000optimization} risk measure has a convex and continuously differentiable formulation, which makes it possible to derive a policy gradient algorithm for optimizing a soft-robust objective that balances between optimizing for expected performance and optimizing for CVaR.
Experiments evaluating PG-BROIL with different prior distributions over reward hypotheses suggest that solving PG-BROIL with different values of $\lambda$ can produce a family of solutions that span the Pareto frontier of policies which trade-off expected performance and robustness. Finally, we show that PG-BROIL improves upon state-of-the-art imitation learning methods when learning from small numbers of demonstrations by not just optimizing for the most likely reward function, but by also hedging against poor performance under other likely reward functions.

\textbf{Future Work and Limitations:} We found that PG-BROIL can sometimes become unstable for values of lambda close to zero---likely due to the indicator function in the CVaR policy gradient. We experimented with entropic risk measure~\cite{follmer2011entropic}, a continuously differentiable alternative to CVaR, but obtained similar results to CVaR (see Appendix~\ref{app:broil_erm_pg_proof}). 
% Additionally, our implementation of PG-BROIL learns a separate value function for each reward function hypothesis to use for advantage estimation 
% Additionally, our experiments assumed access to known reward features. 
%Using reward-conditioned value functions~\cite{schaul2015universal} or learning successor features~\cite{barreto2017successor} could enable more efficient value function estimation 
Future work also includes using contrastive learning~\cite{laskin2020curl} and deep Bayesian reward function inference~\cite{brown2020safe} to enable robust policy learning from raw pixels.
% Additionally, PG-BROIL performs generalized advantage estimation \cite{schulman2015high} and estimates a separate value function network per reward function hypothesis leading to space and time complexity that scales linearly with the number of reward function samples. In the future, techniques such as universal value function approximators~\cite{schaul2015universal} could be used to learn a single reward-conditioned value function approximator to enable highly efficient advantage estimation over a large number of samples. Future work should also investigate using deep learning to learn the reward function features. Our experiments assumed access to known reward function features to enable interpretable results; however, leveraging recent work on efficient deep Bayesian reward function inference has the potential to allow PG-BROIL to scale to more complicated tasks with unknown reward function features \cite{brown2020safe}. 

\section*{Acknowledgements}
% This work was supported in part by NSF SCHooL, AFOSR, ONR YIP, NSF Grants IIS-1717368 and IIS-1815275 and donations from Google, Siemens, Amazon Robotics, Toyota Research Institute, and by equipment  grants  from  NVidia.
The authors would like to thank the anonymous reviewers for their helpful suggestions for improving the paper. This work has taken place in the AUTOLAB and InterACT Lab at the University of California, Berkeley and the Reinforcement Learning and Robustness Lab (RLsquared) at the University of New Hampshire. AUTOLAB research is supported in part by the Scalable Collaborative Human-Robot Learning (SCHooL) Project, the NSF National Robotics Initiative Award 1734633, and in part by donations from Google, Siemens, Amazon Robotics, Toyota Research Institute, and by equipment  grants  from  NVidia. InterACT Lab research is supported in part by AFOSR, NSF NRI SCHOOL, and ONR YIP. RLsquared research is supported in part by NSF Grants IIS-1717368 and IIS-1815275. Ashwin Balakrishna is supported by an NSF GRFP. This  article  solely  reflects  the  opinions  and conclusions of its authors and not the views of the sponsors or their associated entities. 
\bibliography{references}
\bibliographystyle{icml2021}
\appendix
\onecolumn
\section{Full Derivation of CVaR BROIL Policy Gradient} \label{app:broil_cvar_pg_proof}
In this section we derive the complete derivation of the policy gradient objective for BROIL.

\subsection{General Performance Metric}
We will first derive a policy gradient algorithm for any performance metric. Then, we will derive special cases corresponding to particular choices of the performance metric.

We start with the same objective, which we note is a weighted combination of two terms, one of which measures expected performance ($\Ex[\psi(\pi_\theta,R)]$) and the other of which measures tail risk ($\alphacvar \bigg[ \psi(\pi_\theta,R) \bigg]$):
\begin{align} 
\operatorname*{maximize}_{\pi_\theta} \quad &\lambda \cdot \Ex[\psi(\pi_\theta,R)] + (1-\lambda) \cdot \alphacvar \bigg[ \psi(\pi_\theta,R) \bigg]
\end{align}

We want to solve this via a policy gradient algorithm so we need to find the gradient with respect to $\theta$. For the first term we have
\begin{align}
    \nabla_\theta \Ex_{\Pr(R)}[\psi(\pi_\theta,R)]  =&  \Ex_{\Pr(R)}[\nabla_\theta\psi(\pi_\theta,R)] \\
    =& \sum_i \Pr(r_i)\nabla_\theta \psi(\pi_\theta,r_i).
\end{align}

Now consider the gradient of the CVaR term. We have 
\begin{equation}
\nabla_\theta \alphacvar[\psi(\pi_\theta,R)] = \nabla_\theta  \max_\sigma \Bigl(\sigma -\frac{1}{1-\alpha} \sum_i \Pr(r_i)  \big[\sigma - \psi(\pi_\theta,r_i)  \big]_+ \Bigr)
\end{equation}
Here we need to take the gradient with respect to an inner maximization over the auxiliary variable $\sigma$.
To solve for the gradient of this term, first note that given a fixed policy $\pi_\theta$, the objective is piecewise linear in $\sigma$ with switch points at each sample from the posterior ($\psi(\pi_\theta,r_i)$ $\forall r_i$). Thus, we can solve for $\sigma$ via linear programming or just via a line search. If we let $\psi_i = \psi(\pi_\theta,r_i)$ then we can quickly iterate over all reward function hypotheses and solve for $\sigma$ as 
\begin{equation}
\sigma^* = \operatorname*{argmax}_{\sigma \in \{\psi_1,\ldots,\psi_N\}}  \Bigl(\sigma -\frac{1}{1-\alpha} \sum_i \Pr(r_i)  \big[\sigma - \psi_i  \big]_+ \Bigr)
\end{equation}

Given the solution to the above optimization problem, we can now fix $\sigma = \sigma^*$ and then perform a step of policy gradient optimization by following the sub-gradient of CVaR with respect to the policy parameters $\theta$:
\begin{align}
    \nabla_\theta \left( \sigma^* -\frac{1}{1-\alpha} \sum_i \Pr(r_i) \big[\sigma^* - \psi(\pi_\theta,r_i)]  \big]_+ \right) 
    = & -\frac{1}{1-\alpha} \sum_i \Pr(r_i) \nabla_\theta\big[\sigma^* - \psi(\pi_\theta,r_i)  \big]_+ \\
    = &\frac{1}{1-\alpha} \sum_i \Pr(r_i) \bm{1}_{\sigma^* \geq \psi(\pi_\theta,r_i)} \nabla_\theta \psi(\pi_\theta,r_i)
\end{align}
where we use the notation $\bm{1}_x$ to denote the indicator function:
\begin{equation}
    \bm{1}_x = 
    \begin{cases}
    1 &\text{if $x$ is True}\\
    0 & \text{otherwise}
    \end{cases}
\end{equation}

We now can formulate the full BROIL policy gradient update step by blending the policy gradient over the expectation with the policy gradient over the CVaR:
\begin{align} \label{eq:broil_perf_grad}
\nabla_\theta \text{BROIL} =&  \lambda \sum_i \Pr(r_i)\nabla_\theta\psi(\pi_\theta,r_i) +  \frac{1-\lambda}{1-\alpha}  \sum_i \Pr(r_i) \bm{1}_{\sigma^* \geq \psi(\pi_\theta,r_i)} \nabla_\theta\psi(\pi_\theta,r_i)\\
= &  \sum_i \Pr(r_i)\nabla_\theta\psi(\pi_\theta,r_i) \bigg( \lambda +  \frac{1-\lambda}{1-\alpha}  \bm{1}_{\sigma^* \geq \psi(\pi_\theta,r_i)} \bigg)
\end{align}

\subsection{Policy Gradient for Expected Return}

We now consider the case where our performance metric is expected value, i.e., $\psi(\pi_{\theta},R) = \rho(\pi_{\theta},R) = \Ex_{\tau \sim \pi_\theta}[R(\tau)]$. 
Plugging expected value for our performance metric into Equation~\eqref{eq:broil_perf_grad} gives the following:
\begin{align}
\nabla_\theta \text{BROIL} =  \sum_i \Pr(r_i)\nabla_\theta \rho(\pi,r_i) \bigg( \lambda +  \frac{1-\lambda}{1-\alpha}  \bm{1}_{\sigma^* \geq \rho(\pi,r_i)} \bigg),
\end{align}
where
solving for $\sigma^*$ requires estimating $\rho_i$ by collecting a set $\mathcal{T}$ of on-policy trajectories $\tau \sim \pi_\theta$ where $\tau = (s_0,a_0,s_1,a_1,\ldots,s_T,a_T)$:
\begin{equation} \label{app-eq:exp_return_hyps}
    \rho_i \approx\frac{1}{|\mathcal{T}|} \sum_{\tau \in \mathcal{T}} \sum_{t=0}^T r_i(s_t, a_t).
\end{equation}
Given the expected return under each reward function hypothesis we solve for $\sigma^*$ as 
\begin{equation}
\sigma^* = \operatorname*{argmax}_{\sigma \in \{\rho_1,\ldots,\rho_N\}}  \Bigl(\sigma -\frac{1}{1-\alpha} \sum_{i=1}^{N} \Pr(r_i)  \big[\sigma - \rho_i  \big]_+ \Bigr).
\end{equation}

Solving for $\sigma^*$ does not require additional data collection beyond what is required for standard policy gradient approaches. We simply evaluate the set of rollouts $\mathcal{T}$ from $\pi_{\theta}$ under each reward function hypothesis, $r_i$ and then solve the optimization problem above to find $\sigma^*$. While this requires more computation than a standard policy gradient approach---we have to evaluate each rollout under $N$ reward functions---this does not increase the online data collection, which is often the bottleneck in RL algorithms.%, adding more offline 

Note that, in general, we can write the policy gradient of the expected return as
\begin{equation}
    \nabla_\theta \rho(\pi,r_i) = \nabla_\theta \Ex_{\tau \sim \pi_\theta} [ r_i(\tau)] = \Ex_{\tau \sim \pi_\theta} \left[ \sum_{t=0}^T \nabla_\theta \log \pi_\theta(a_t \mid s_t) \Phi^{r_i}_t \right]
\end{equation}
where $\Phi^{r_i}_t$ is some measure of the quality of the policy under reward function $r_i$. Common choices include the return of a trajectory:
%\begin{equation}
    $\Phi^{r_i}_t = r_i(\tau)$,
%\end{equation}
the reward-to-go from time $t$:
%\begin{equation}
    $\sum_{t'=t}^T r_i(s_{t'}, a_{t'})$,
%\end{equation}
the reward-to-go with a state-dependent baseline:
%\begin{equation}
    $\sum_{t'=t}^T r_i(s_{t'}, a_{t'}) - b(s_t)$,
%\end{equation}
the on-policy action-value function
%\begin{equation}
     $Q^{\pi_\theta}(s_t, a_t)$,
%\end{equation}
or the on-policy advantage function (the most popular choice) \cite{schulman2015high}:
\begin{equation}
    \Phi^{r_i}_t = A^{\pi_\theta}(s_t, a_t) = Q^{\pi_\theta}(s_t, a_t) - V^{\pi_\theta}(s_t).
\end{equation}

Any of these formulations of the policy gradient can be used for the above BROIL policy gradient as follows where we approximate the expectation using a set $\mathcal{T}$ of on-policy trajectories $\tau \sim \pi_\theta$:

\begin{align}
\nabla_\theta \text{BROIL} =& \sum_i \Pr(r_i)\nabla_\theta\Ex_{\tau \sim \pi_\theta}[r_i(\tau)] \bigg( \lambda +  \frac{1-\lambda}{1-\alpha}  \bm{1}_{\sigma^* \geq \rho(\pi,r_i)} \bigg) \\
=& \sum_i \Pr(r_i) \bigg(\Ex_{\tau \sim \pi_\theta} \left[ \sum_{t=0}^T \nabla_\theta \log \pi_\theta(a_t \mid s_t) \Phi^{r_i}_t \right] \bigg) \bigg( \lambda +  \frac{1-\lambda}{1-\alpha}  \bm{1}_{\sigma^* \geq \rho(\pi,r_i)} \bigg) \\
\approx& \sum_i \Pr(r_i) \bigg(\frac{1}{|\mathcal{T}|} \sum_{\tau \in \mathcal{T}} \left[ \sum_{t=0}^T \nabla_\theta \log \pi_\theta(a_t \mid s_t) \Phi^{r_i}_t \right] \bigg) \bigg( \lambda +  \frac{1-\lambda}{1-\alpha}  \bm{1}_{\sigma^* \geq \rho(\pi,r_i)} \bigg) \\
=& \frac{1}{|\mathcal{T}|}\sum_i \Pr(r_i) \bigg( \sum_{\tau \in \mathcal{T}} \left[ \sum_{t=0}^T \nabla_\theta \log \pi_\theta(a_t \mid s_t) \Phi^{r_i}_t \right] \bigg) \bigg( \lambda +  \frac{1-\lambda}{1-\alpha}  \bm{1}_{\sigma^* \geq \rho(\pi,r_i)} \bigg) \\
=& \frac{1}{|\mathcal{T}|}\sum_i  \sum_{\tau \in \mathcal{T}} \Pr(r_i) \left[ \sum_{t=0}^T \nabla_\theta \log \pi_\theta(a_t \mid s_t) \Phi^{r_i}_t \right] \bigg( \lambda +  \frac{1-\lambda}{1-\alpha}  \bm{1}_{\sigma^* \geq \rho(\pi,r_i)} \bigg) \\
=& \frac{1}{|\mathcal{T}|} \sum_{\tau \in \mathcal{T}}\sum_i \Pr(r_i) \left[ \sum_{t=0}^T \nabla_\theta \log \pi_\theta(a_t \mid s_t) \Phi^{r_i}_t \right] \bigg( \lambda +  \frac{1-\lambda}{1-\alpha}  \bm{1}_{\sigma^* \geq \rho(\pi,r_i)} \bigg) \\
=& \frac{1}{|\mathcal{T}|} \sum_{\tau \in \mathcal{T}}\sum_i   \sum_{t=0}^T \Pr(r_i)\nabla_\theta \log \pi_\theta(a_t \mid s_t) \Phi^{r_i}_t  \bigg( \lambda +  \frac{1-\lambda}{1-\alpha}  \bm{1}_{\sigma^* \geq \rho(\pi,r_i)} \bigg) \\
=& \frac{1}{|\mathcal{T}|} \sum_{\tau \in \mathcal{T}}   \sum_{t=0}^T \sum_i \Pr(r_i)\nabla_\theta \log \pi_\theta(a_t \mid s_t) \Phi^{r_i}_t  \bigg( \lambda +  \frac{1-\lambda}{1-\alpha}  \bm{1}_{\sigma^* \geq \rho(\pi,r_i)} \bigg) \\
=& \frac{1}{|\mathcal{T}|} \sum_{\tau \in \mathcal{T}}  \sum_{t=0}^T \nabla_\theta \log \pi_\theta(a_t \mid s_t) \bigg(\sum_i \Pr(r_i)\Phi_t^{r_i}(\tau)\big(\lambda   +  \frac{1-\lambda}{1-\alpha} \bm{1}_{\sigma^* \geq \rho(\pi,r_i)} \big) \bigg) \\
=& \frac{1}{|\mathcal{T}|} \sum_{\tau \in \mathcal{T}}  \sum_{t=0}^T \nabla_\theta \log \pi_\theta(a_t \mid s_t) w_t 
\end{align}
where 
\begin{equation}w_t = \sum_i \Pr(r_i)\Phi_t^{r_i}(\tau)\left(\lambda   +  \frac{1-\lambda}{1-\alpha} \bm{1}_{\sigma^* \geq \rho(\pi,r_i)} \right)
\end{equation}
is the weight associated with each state-action pair. Intuitively, if $\lambda=1$, then we just focus on increasing the likelihood of actions that look good in expectation. If $\lambda=0$, then we focus on increasing the likelihood of actions that look good under reward functions that the current policy $\pi_\theta$ performs poorly under, i.e., we focus on improving our performance under all $r_i$ such that $\sigma^* > \rho(\pi,r_i)$), weighting the gradient according to the likelihood of these worst-case reward functions.

\subsection{Policy Gradient for Baseline Regret}
We now consider the case where our performance metric is baseline regret~\cite{brown2020bayesian}, which measures performance with respect to some expert demonstrator. The intuition is that this formulation may be able to reduce variance in the policy gradient estimator by grounding updates in the expected return of the demonstrator. We define baseline regret as follows:
\begin{equation}
    \psi(\pi_{\theta},R) = \rho(\pi_\theta,R) - \rho(\pi_{E},R),
\end{equation}
where $\pi_E$ denotes an expert policy and $\rho(\pi_{E},R)$ is usually estimated from demonstrations. Plugging baseline regret for our performance metric into Equation~\eqref{eq:broil_perf_grad} gives the following:
\begin{align}
\nabla_\theta \text{BROIL} &=  \sum_i \Pr(r_i)\nabla_\theta \big( \rho(\pi_\theta,r_i) - \rho(\pi_{E},r_i) \big) \bigg( \lambda +  \frac{1-\lambda}{1-\alpha}  \bm{1}_{\sigma^* \geq \rho(\pi_\theta,r_i) - \rho(\pi_{E},r_i)} \bigg)\\
&=  \sum_i \Pr(r_i)\nabla_\theta  \rho(\pi_\theta,r_i) \bigg( \lambda +  \frac{1-\lambda}{1-\alpha}  \bm{1}_{\sigma^* \geq \rho(\pi_\theta,r_i) - \rho(\pi_{E},r_i)} \bigg)\\
&= \sum_i \Pr(r_i)\nabla_\theta  \Ex_{\tau \sim \pi_\theta}[r_i(\tau)] \bigg( \lambda +  \frac{1-\lambda}{1-\alpha}  \bm{1}_{\sigma^* \geq \rho(\pi_\theta,r_i) - \rho(\pi_{E},r_i)} \bigg)
\end{align}

In practice, we typically only have samples of expert behavior rather than a full policy. In this case, we can estimate the return of the demonstrator under reward function hypothesis $r_i$ using a set of demonstrated trajectories $D = \{\tau_1, \ldots, \tau_m \}$ as
\begin{equation}
        \rho(\pi_E, r_i) \approx\frac{1}{|\mathcal{D}|} \sum_{\tau \in \mathcal{D}} \sum_{t=0}^T r_i(s_t, a_t),
\end{equation}
where $T$ is the horizon of the demonstrations.

If $r_i$ is a linear function, i.e.,$(s,a) = \wb_i^T \phi(s,a)$, then we can compute the empirical expected feature counts using the demonstrated trajectories $D = \{\tau_1, \ldots, \tau_m \}$ to get 
\begin{equation}
    \hat{\mu}_E = \frac{1}{|\mathcal{D}|} \sum_{\tau \in \mathcal{D}} \sum_{(s_t,a_t) \in \tau} \phi(s_t,a_t), 
\end{equation}
where $\phi:\mathcal{S}\times \mathcal{A} \to \Real^k$ denotes the reward features. We can then estimate $\rho(\pi_E,r_i)$ as
\begin{equation}
    \rho(\pi_E, r_i) =  \wb_i^T \hat{\mu}_E,
    \end{equation}
where $\bm{w}_i$ is  the feature weight vector corresponding to linear reward function $r_i$ sampled from the posterior. The advantage is that we only have to evaluate the expected feature counts once and then we can use this vector to estimate the expected return under any number of reward function hypotheses via dot products.

Given the estimate baseline regret under each reward function hypothesis we solve for $\sigma^*$ as 
\begin{equation} \label{app-eq:solve_for_sigma}
\sigma^* = \operatorname*{argmax}_{\sigma \in \{\rho^{\rm br}_1,\ldots,\rho^{\rm br}_N\}}  \Bigl(\sigma -\frac{1}{1-\alpha} \sum_{i=1}^{N} \Pr(r_i)  \big[\sigma - \rho^{\rm br}_i  \big]_+ \Bigr),
\end{equation}
where $\rho^{\rm br}_i = \rho(\pi_\theta,r_i) - \rho(\pi_{E},r_i)$.

As in the previous section, if we approximate the baseline regret using a set $\mathcal{T}$ of on-policy trajectories $\tau. \sim \pi_\theta$ and a set $\mathcal{D}$ of demonstrations we have:
\begin{align}
\nabla_\theta \text{BROIL} =& \sum_i \Pr(r_i)\nabla_\theta\Ex_{\tau \sim \pi_\theta}[r_i(\tau)] \bigg( \lambda +  \frac{1-\lambda}{1-\alpha}  \bm{1}_{\sigma^* \geq \rho^{\rm br}_i} \bigg) \\
=& \sum_i \Pr(r_i) \bigg(\Ex_{\tau \sim \pi_\theta} \left[ \sum_{t=0}^T \nabla_\theta \log \pi_\theta(a_t \mid s_t) \Phi^{r_i}_t \right] \bigg) \bigg( \lambda +  \frac{1-\lambda}{1-\alpha}  \bm{1}_{\sigma^* \geq \rho^{\rm br}_i} \bigg) \\
\approx& \sum_i \Pr(r_i) \bigg(\frac{1}{|\mathcal{T}|} \sum_{\tau \in \mathcal{T}} \left[ \sum_{t=0}^T \nabla_\theta \log \pi_\theta(a_t \mid s_t) \Phi^{r_i}_t \right] \bigg) \bigg( \lambda +  \frac{1-\lambda}{1-\alpha}  \bm{1}_{\sigma^* \geq \rho^{\rm br}_i} \bigg) \\
=& \frac{1}{|\mathcal{T}|} \sum_{\tau \in \mathcal{T}}  \sum_{t=0}^T \nabla_\theta \log \pi_\theta(a_t \mid s_t) w_t 
\end{align}
where 
\begin{equation}w_t = \sum_i \Pr(r_i)\Phi_t^{r_i}(\tau)\big(\lambda   +  \frac{1-\lambda}{1-\alpha} \bm{1}_{\sigma^* \geq \rho^{\rm br}_i} \big)
\end{equation}
is the weight associated with each state-action pair. The baseline regret adjusts risk such that it is riskier to explore areas of the state-space that were not visited by the demonstrator, thereby encouraging pessimism in the face of uncertainty. To see this note that 
\begin{equation}
    \rho^{\rm br}_i = \rho(\pi_\theta,r_i) - \rho(\pi_{E},r_i) \approx  \wb_i^T (\hat{\mu}_{\pi_\theta} - \hat{\mu}_E) = \sum_{j=1}^k \wb_i[j] (\hat{\mu}_{\pi_\theta}[j] - \hat{\mu}_E[j]),
\end{equation}
where $\hat{\mu}_{\pi_\theta}$ are the estimated expected feature counts of $\pi_\theta$ and  $\hat{\mu}_E$ are the estimated expected feature counts of $\pi_E$ and we assume all vectors lie in $\mathbb{R}^k$. Thus, if the expert and policy both encounter reward feature $j$ at the same frequency ($\hat{\mu}_{\pi_\theta}[j] = \hat{\mu}_E[j]$), the distribution over $\wb_i[j]$ will not contribute to $\rho^{\rm br}_i$. Thus, the tail risk will be determined by other reward weight distributions. Conversely, when there is disagreement, there will be the potential for risk: if the policy visits new states that are estimated to have negative reward weight or if the policy does not visit states visited by the demonstrator that are estimated to have positive reward weight, then either will lower $\rho^{\rm br}_i$ and result in more tail risk.

Note, however, that baseline regret does not only provide an incentive to directly imitate the demonstrator. If demonstrations are suboptimal, but we have preferences over them, \citep{brown2020safe} demonstrated that fast Bayesian reward inference is possible. If under the posterior distribution of reward functions we have high confidence that certain states are good (positive weight) or bad (negative weight), then lower risk policies will seek to visit the bad states less often than the demonstrator and visit the good states more often. Thus, it is still possible to outperform the demonstrator while being robust to reward weights with high uncertainty by imitating to hedge against high uncertainty, but exploiting our posterior to perform better than the demonstrator when we have low uncertainty over the desirability of certain states.
\section{Entropic Risk Measure Policy Gradient}\label{app:broil_erm_pg_proof}
Here we show that another common risk metric, Entropic Risk Measure (ERM) \cite{follmer2011entropic}, also is amenable to policy gradient optimization within the BROIL framework. One benefit of ERM is that it is differentiable everywhere unlike CVaR. ERM has been considered recently under the settings of risk-averse policy search under a known reward function~\cite{nass2019entropic} and soft-robust optimization with respect to model uncertainty~\cite{russel2020entropic}.

\subsection{Entropic Risk Measure}
The entropic risk measure \cite{follmer2011entropic} is another form of tail risk that has the benefit of being everywhere differentiable. The entropic risk measure (ERM) of a random variable $X$ is defined as:
\begin{equation}
    ERM = -\frac{1}{\alpha} \log \Ex [ e^{-\alpha X}]
\end{equation}
where $\alpha \in (0,\infty)$ represents the risk sensitivity (higher is more risk-sensitive) and where larger values of ERM indicate lower risk.

Similar to the CVaR BROIL objective we can formulate at BROIL objective using ERM. As we show in Section~\ref{app:ERM-deriv}, the policy gradient of ERM-BROIL is given by Equation~\ref{eq:pg_broil_general} with 
\begin{equation}
    w^{ERM}_t = \sum_i \Pr(r_i) \Phi^{r_i}_t(\tau) \left(\lambda + (1-\lambda) \frac{e^{-\alpha \rho(\pi_\theta, r_i)}}{\Ex_R [e^{-\alpha \rho(\pi_\theta, R)}]} \right)
\end{equation}

If $\lambda=1$, then we just focus on increasing the likelihood of actions that look good in expectation. If $\lambda=0$, then we focus on increasing the likelihood of actions that look good under reward functions that the current policy $\pi_\theta$ performs poorly under. In particular, the policy gradient for the ERM term is given by a weighted sum of policy gradients for each reward function in the posterior. The weights are softmax probabilities which will concentrate the probability around the reward function $r_i$ for which $\rho(\pi_\theta, r_i)$ is lowest. Intuitively, this will encourage policy updates that improve the performance under the reward functions for which $\pi_\theta$ performs the worst. As $\alpha \rightarrow \infty$, the softmax probabilities will concentrate on the absolute worst-case reward in the distribution, but for $\alpha \rightarrow 0$, this probability will be distributed according to the reward function probabilities $\Pr(r_i)$ resulting in a policy gradient that seeks to maximize return under the expected reward function. 

\subsection{Deriviation}
\label{app:ERM-deriv}

In this section we derive a similar policy gradient objective for BROIL that uses entropic risk measure:
\begin{equation}
    \text{ERM}_\alpha = -\frac{1}{\alpha} \log( \Ex_R[e^{-\alpha \psi(\pi_\theta,R)}])
\end{equation}

We start with the objective:
\begin{align} 
\operatorname*{maximize}_{\pi_\theta} \quad &\lambda \cdot \Ex[\psi(\pi_\theta,R)] + (1-\lambda) \cdot \text{ERM}_\alpha \bigg[ \psi(\pi_\theta,R) \bigg]
\end{align}

We assume that our performance metric is expected value, i.e., $\bm{\psi}(\pi_{\bm{u}},R) = \rho(\pi,R) = \Ex_{\tau \sim \pi_\theta}[R(\tau)]$.

We need to find the gradient wrt $\theta$. The first term is the same as in the previous section:
\begin{align}
    \nabla_\theta \cdot \Ex_{\Pr(R)}[\Ex_{\tau \sim \pi_\theta}[R(\tau)]] 
    =& \sum_i \Pr(r_i)\nabla_\theta\Ex_{\tau \sim \pi_\theta}[r_i(\tau)].
\end{align}

Now consider the gradient of the entropic risk term. We have 
\begin{align}
\nabla_\theta \text{ERM}_\alpha[\rho(\pi,R)] =& -\nabla_\theta  \frac{1}{\alpha} \log\left(\sum_i \Pr(r_i)  e^{-\alpha \rho(\pi_\theta, r_i)} \right) \\
=&-\frac{1}{\alpha} \frac{1}{\sum_j \Pr(R_j)  e^{-\alpha \rho(\pi_\theta, R_j)}}   \sum_i \Pr(r_i)  \nabla_\theta e^{-\alpha \rho(\pi_\theta, r_i)} \\
=&-\frac{1}{\alpha} \frac{1}{\sum_j \Pr(R_j)  e^{-\alpha \rho(\pi_\theta, R_j)}}   \sum_i \Pr(r_i)  e^{-\alpha \rho(\pi_\theta, r_i)} \nabla_\theta  (-\alpha \rho(\pi_\theta, r_i)) \\
=& \sum_i \frac{\Pr(r_i) e^{-\alpha \rho(\pi_\theta, r_i)}}{\sum_j \Pr(R_j)  e^{-\alpha \rho(\pi_\theta, R_j)}}       \nabla_\theta  \rho(\pi_\theta, r_i)
\end{align}

% The policy gradient is given by a weighted sum of policy gradients for each reward function in the posterior. The weights are softmax probabilities which will concentrate the probability around the reward function $r_i$ for which $\rho(\pi_\theta, r_i)$ is lowest. Intutively, this will encourage policy updates that improve the performance under the reward functions for which $\pi_\theta$ performs the worst. As $\alpha \rightarrow \infty$, the softmax probability will concentrate on the absolute worst-case reward in the distribution, but for $\alpha \rightarrow 0$, this probability will be distributed according to the reward function probabilities $\Pr(r_i)$ resulting in a policy gradient that seeks to maximize return under the expected reward function. 

As before we will be estimating the on-policy expected return for each reward hypothesis which can be done by collecting a set $\mathcal{T}$ of trajectories $\tau \sim \pi_\theta$:
\begin{equation}
    \rho(\pi_\theta,R_j) = \Ex_{\tau \sim \pi_\theta}[r_ij(\tau)] \approx \frac{1}{|\mathcal{T}|} \sum_{\tau \in \mathcal{T}} R_j(\tau) = \frac{1}{|\mathcal{T}|} \sum_{\tau \in \mathcal{T}} \sum_{t=0}^T R_j(s_t, a_t).
\end{equation}

Now we can formulate the full BROIL policy gradient update step by blending the policy gradient over the expectation with the policy gradient over the ERM:
\begin{align}
\nabla_\theta \text{BROIL} =&  \lambda \sum_i \Pr(r_i)\nabla_\theta\rho(\pi_\theta, r_i) +  (1-\lambda)\sum_i \frac{\Pr(r_i) e^{-\alpha \rho(\pi_\theta, r_i)}}{\sum_j \Pr(R_j)  e^{-\alpha \rho(\pi_\theta, R_j)}}       \nabla_\theta  \rho(\pi_\theta, r_i)\\
= &  \sum_i \Pr(r_i)\nabla_\theta  \rho(\pi_\theta, r_i) \bigg( \lambda +  (1-\lambda) \frac{ e^{-\alpha \rho(\pi_\theta, r_i)}}{\Ex_{\Pr(R)}[ e^{-\alpha \rho(\pi_\theta, R)}]}        \bigg)
\end{align}

As before we can write the policy gradient as
\begin{equation}
    \nabla_\theta  \rho(\pi_\theta, r_i) = \nabla_\theta \Ex_{\tau \sim \pi_\theta} [ r_i(\tau)] = \Ex_{\tau \sim \pi_\theta} \left[ \sum_{t=0}^T \nabla_\theta \log \pi_\theta(a_t \mid s_t) \Phi^{r_i}_t \right].
\end{equation}

Defining $\Phi^{r_i}_t$ in terms of a particular reward function hypothesis $r_i$ and approximating expectations with a set $\mathcal{T}$ of on-policy trajectories $\tau \sim \pi_\theta$ gives:
\begin{align}
\nabla_\theta \text{BROIL} 
\approx& \sum_i \Pr(r_i) \bigg(\frac{1}{|\mathcal{T}|} \sum_{\tau \in \mathcal{T}} \left[ \sum_{t=0}^T \nabla_\theta \log \pi_\theta(a_t \mid s_t) \Phi^{r_i}_t \right] \bigg) \bigg( \lambda +  (1-\lambda)  \frac{ e^{-\alpha \rho(\pi_\theta, r_i)}}{\Ex_R[ e^{-\alpha \rho(\pi_\theta, R)}]}        \bigg) \\
=& \frac{1}{|\mathcal{T}|} \sum_{\tau \in \mathcal{T}}   \sum_{t=0}^T \sum_i \Pr(r_i)\nabla_\theta \log \pi_\theta(a_t \mid s_t) \Phi^{r_i}_t  \bigg( \lambda +  (1-\lambda)  \frac{ e^{-\alpha \rho(\pi_\theta, r_i)}}{\Ex_R[ e^{-\alpha \rho(\pi_\theta, R)}]}        \bigg) \\
=& \frac{1}{|\mathcal{T}|} \sum_{\tau \in \mathcal{T}}  \sum_{t=0}^T \nabla_\theta \log \pi_\theta(a_t \mid s_t) \bigg(\sum_i \Pr(r_i)\Phi_t^{r_i}(\tau)\bigg( \lambda +  (1-\lambda) \frac{ e^{-\alpha \rho(\pi_\theta, r_i)}}{\Ex_R[ e^{-\alpha \rho(\pi_\theta, R)}]}        \bigg) \bigg) \\
=& \frac{1}{|\mathcal{T}|} \sum_{\tau \in \mathcal{T}}  \sum_{t=0}^T \nabla_\theta \log \pi_\theta(a_t \mid s_t) w_t 
\end{align}
where 
\begin{equation}
w_t = \sum_i \Pr(r_i)\Phi_t^{r_i}(\tau)\bigg( \lambda +  (1-\lambda) \frac{ e^{-\alpha \rho(\pi_\theta, r_i)}}{\Ex_R[ e^{-\alpha \rho(\pi_\theta, R)}]}        \bigg)
\end{equation} is the weight associated with each state-action pair. Intuitively, if $\lambda=1$, then we just focus on increasing the likelihood of actions that look good in expectation. If $\lambda=0$, then we focus on increasing the likelihood of actions that look good under reward functions that the current policy $\pi_\theta$ performs poorly under.

\subsection{Experiments}
\begin{figure}[h]
    \centering
    \includegraphics[width=.4\linewidth]{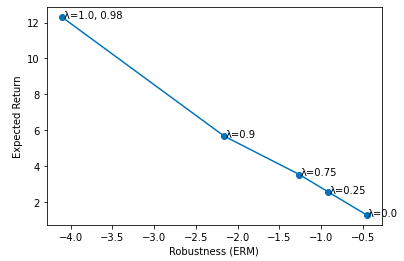}
    \caption{Efficient frontier curve for CartPole with the ERM risk metric. We set $\alpha$ equal to 0.001 and test over multiple values of $\lambda$. PG-BROIL with ERM acheives similar stability to CVaR.}
    \label{fig:erm_cartpole_frontier}
\end{figure}

\paragraph{CartPole}
Using the same posterior and same hyperparameters as the original experiment, we redo the experiment except using ERM as the risk metric. Figure \ref{fig:erm_cartpole_frontier} shows the tradeoff between robustness and expected return for various $\lambda$. We find that results with the ERM risk metric are relatively similar to those with the CVaR risk metric in the main text.

\begin{figure}[!ht]
\centering
\includegraphics[width=.49\textwidth]{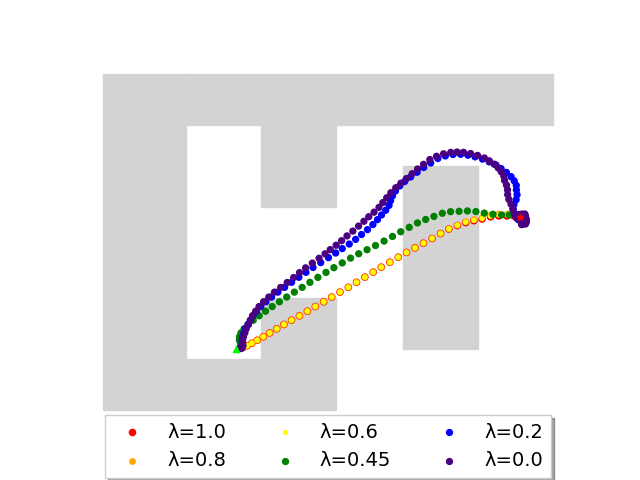}
\includegraphics[width=.42\textwidth]{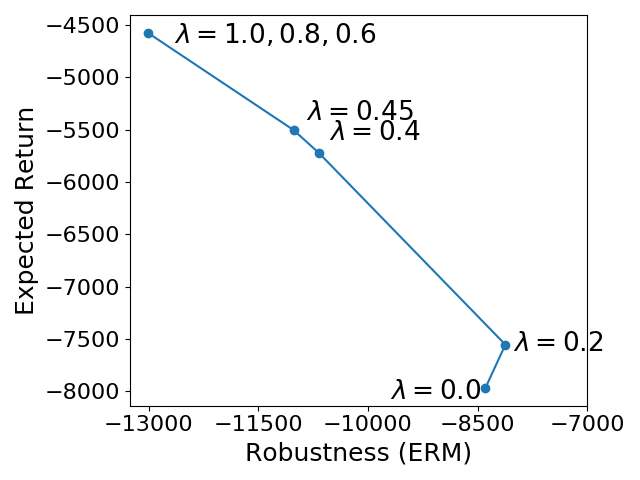}
\caption{We show qualitative performance (left) and an efficient frontier curve (right) for the same environment and parameters as Figure~\ref{fig:reward-priors-domains}b, but use ERM as the risk measure instead of CVaR for different values of lambda.}
\label{app:pb}
\end{figure}

\paragraph{Pointmass Navigation}
Figure \ref{app:pb} shows the Pointmass Navigation task with the ERM risk measure. Overall, the behavior is very similar. One distinction is that for lower values of lambda (ie $0, 0.2$) the pointmass goes through the edge of the gray region while for CVaR the pointmass avoided the gray region entirely.

\paragraph{TrashBot}

\begin{figure}
\centering
\includegraphics[width=.45\linewidth]{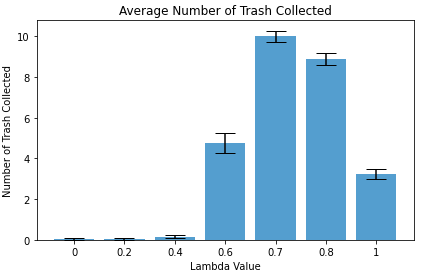}
\includegraphics[width=.45\linewidth]{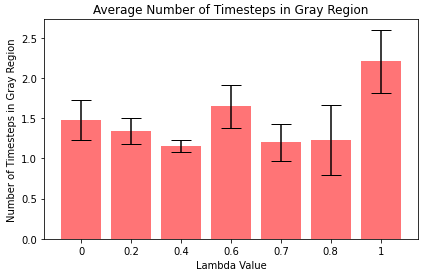}
\caption{We run the TrashBot environment with the ERM risk metric and $\alpha = 1$ over various values of $\lambda$. 
%and find that as expected, the TrashBot collects more trash when $\lambda$ is higher as it more heavily weights expected performance and does not focus excessively on tail risk. 
We take the 95\% confidence interval and plot them as the error bars. 
%Additionally, as expected, as $\lambda$ is increased the TrashBot is less risk sensitive, and thus spends more time in the gray region.
We find that the TrashBot collects the most trash when $\lambda = 0.7$ while minimizing the amount of time in the grey region.
}
\label{fig:erm_trashbot}
\end{figure}

Figure \ref{fig:erm_trashbot} shows the same TrashBot experiment with the ERM risk metric and $\alpha=1$. We find that results are similar when ERM is used instead of CVaR. With CVaR we saw $\lambda=0.8$ gave the best results while for ERM $\lambda=0.7$ was best.

\vspace{3cm}

\section{Trust Region PG-BROIL}\label{app:ppo_broil}

We now derive a version of the Proximal Policy Optimization (PPO)~\cite{schulman2017proximal} algorithm for optimizing the BROIL objective. We specifically consider the PPO-clip objective, which adjusts the advantage function to encourage controlled updates of the policy at each epoch. Precisely, let the policy parameters at epoch $k$ be given by $\theta_{k}$. Then PPO-clip implements the following update:
\begin{equation}
    \theta_{k+1} = \operatorname*{argmax}_\theta \Ex_{(s, a) \sim \pi_{\theta_k}} [ L(a, s, \theta_{k}, \theta)]
\end{equation}
where
\begin{equation}
    L(a, s, \theta_{k}, \theta) = \min\left(\frac{\pi_\theta (a|s)}{\pi_{\theta_k} (a|s)} A^{\pi_{\theta_k}}(s, a), 
    g(\epsilon, A^{\pi_{\theta_k}} (s, a))\right)
\end{equation}
and
\begin{equation}
    g(\epsilon, A^{\pi_{\theta_k}}(s, a)) = \begin{cases}
    (1+\epsilon) A^{\pi_{\theta_k}}(s, a) & \text { $A^{\pi_{\theta_k}}(s, a) \geq 0$} \\
    (1-\epsilon) A^{\pi_{\theta_k}}(s, a) & \text { $A^{\pi_{\theta_k}}(s, a) < 0$}
    \end{cases}
\end{equation}

% Here ${\pi_\theta}$ denotes a policy with parameters ${\theta}$ and ${{\pi_\theta}_{k}}$ denotes the old policy with parameters ${\theta_{k}}$. $\frac{\pi_\theta (a|s)}{{\pi_\theta}_{k} (a|s)} A^{{\pi_\theta}_{k}} (s, a)$ is a measure of how policy ${\pi_\theta}$ performs relative to the old policy ${{\pi_\theta}_{k}}$ using data from the old policy. $\epsilon$ denotes the clip ratio, or how far away the new policy can go from the old policy, so that $g(\epsilon, A^{{\pi_\theta}_{k}} (s, a))$ serves as a regularizer that puts a limit on how much the new policy can diverge from the old policy. 

To implement a PPO-style gradient clipping for PG-BROIL, we replace $A^{{\pi_\theta}_{k}} (s, a)$ with the BROIL Policy Gradient weights: 

\begin{equation} \label{eq:broil_pg_weight_1}
w_t = \sum_i \Pr(r_i)\Phi_t^{r_i}(\tau)\big(\lambda   +  \frac{1-\lambda}{1-\alpha} \bm{1}_{\sigma^* \geq \rho(\pi,r_i)} \big)
\end{equation}

where $w_t$ is the weight associated with each state-action pair. 

% $\Phi_t$ can be the return of a trajectory
% %\begin{equation}
%     $\Phi_t = R(\tau)$,
% %\end{equation}
% or the reward-to-go from time $t$
% %\begin{equation}
%     $\Phi_t = \sum_{t'=t}^T R(s_{t'}, a_{t'})$,
% %\end{equation}
% or the reward-to-go minus a baseline that only depends on states
% %\begin{equation}
%     $\Phi_t = \sum_{t'=t}^T R(s_{t'}, a_{t'}) - b(s_t)$,
% %\end{equation}
%  or the on-policy action-value function
% %\begin{equation}
%     $\Phi_t = Q^{\pi_\theta}(s_t, a_t)$,
% %\end{equation}
% or most commonly as the on-policy advantage function
% %\begin{equation}
%     $\Phi_t = A^{\pi_\theta}(s_t, a_t) = Q^{\pi_\theta}(s_t, a_t) - V^{\pi_\theta}(s_t)$.
% %\end{equation}

The full PPO-clip objective for BROIL is shown in Algorithm~\ref{alg:ppo_broil}.
\begin{algorithm}
   \caption{PPO-clip BROIL}
   \label{alg:ppo_broil}
    \begin{algorithmic}[1]
    
     \STATE {\bfseries Input:} initial policy parameters $\theta_0$, samples from reward function posterior $R_1,\ldots,R_N$ and associated probabilities, $\Pr(R_1),\ldots,\Pr(R_N)$, and any form for policy gradient weights $\Phi_t$

        \FOR{$k=0,1,2,\ldots$}
            \STATE Collect set of trajectories $\mathcal{T}_k = \{ \tau_i \}$ by running policy $\pi_{\theta}$ in the environment.
            \STATE Estimate expected return of $\pi_{\theta}$ under each reward function hypothesis $r_j$ using~Eq.~\eqref{eq:exp_return_hyps}.
            
            %\begin{equation*}
            %\hat{\rho}_i = \frac{1}{|\mathcal{T}|} \sum_{\tau \in \mathcal{T}} \sum_{t=0}^T r_i(s_t, a_t).
            % \end{equation*}

            \STATE Solve for $\sigma^*$ using Eq.~\eqref{eq:solve_for_sigma}
            
            % \begin{equation*}
            % \sigma^* = \operatorname*{argmax}_{\sigma \in \{\hat{\rho}_1, \ldots, \hat{\rho}_N \}}  \Bigl(\sigma -\frac{1}{1-\alpha} \sum_i \Pr(r_i)  \big[\sigma - \hat{\rho}_i  \big]_+ \Bigr)
            % \end{equation*}
            
            \STATE Update $\theta$ with stochastic gradient ascent by maximizing the PPO-clip objective:
        
            \begin{equation*}
                \theta_{k+1} = \operatorname*{argmax}_\theta \frac{1}{|\mathcal{T}|} \sum_{\tau \in \mathcal{T}}  \biggl[\frac{1}{T} \sum_{t=0}^T \operatorname*{min} \left( \frac{\pi_\theta (a|s)}{{\pi_\theta}_{k} (a|s)} {w_t}, 
                g(\epsilon,  w_t) \right) \biggr]
            \end{equation*}

            using Eq.~\eqref{eq:broil_pg_weight_1} for $w_t$.

            % \begin{equation*}
            %     \hat{g}_k = \frac{1}{|\mathcal{T}|} \sum_{\tau \in \mathcal{T}} \bigg[ \sum_{t=0}^T \nabla_\theta \log \pi_{\theta_k}(a_t \mid s_t) w_t \bigg]
            % \end{equation*}
            % where $w_t = \sum_i \Pr(r_i)\Phi_t^{r_i}(\tau)\big(\lambda   +  \frac{1-\lambda}{1-\alpha} \bm{1}_{\sigma^* > \hat{\rho}_i)} \big)$.
            
        \ENDFOR
    \end{algorithmic}
\end{algorithm}

\section{Experiment Hyperparameters and Details}
The hyperparameters used for PPO are in Table \ref{app:hyperparams}, unless otherwise specified in the experiment's individual section.

\begin{table}[b]
\caption{PG-BROIL hyperparameters when built on PPO.}
\label{app:hyperparams}
\vskip 0.15in
\begin{center}
\begin{small}
\begin{sc}
\scalebox{0.9}{\begin{tabular}{ll}
\toprule
Hyperparameter & \begin{tabular}[c]{@{}l@{}}Value \end{tabular} \\
\midrule
Clip Ratio & 0.2 \\
Environment Steps per Epoch & 4000 \\
GAE Lambda & 0.95 \\
Gamma & 0.99 \\
Hidden Units & 64 \\
Network Layers & 2 \\
Optimizer & Adam \\
Policy Learning Rate & 2e-4\\
Target KL & 0.01\\
Value Learning Rate & 1e-3\\
\bottomrule
\end{tabular}}
\end{sc}
\end{small}
\end{center}
\vskip -0.1in
\end{table}

\subsection{Cart Pole}\label{app:cartpole_details}
We modify the Open AI Gym Cartpole environment~\cite{gym} but modify the reward function to be a linear function of the cart's position by taking the cart position and multiplying it by -1, -0.8, -0.6, -0.4, -0.2, 0, and 0.2 to get our multiple reward hypotheses. For policy optimization, we implement PG-BROIL on top of the REINFORCE implementation from~\cite{spinning-up} with all parameters set to their default settings except for $\alpha = 0.95$ and epochs set to 100.

\subsection{Pointmass Navigation}
\label{app:pointbot_details}
We build on the pointmass navigation environment from~\cite{SAVED} and construct a system in which a pointmass agent navigates from a fixed start state to a fixed goal state with linear Gaussian dynamics. The agent can exert force in cardinal directions and experiences drag coefficient $\psi$ and Gaussian process noise $z_t \sim \mathcal{N}(0, \sigma^2 I)$ in the dynamics. We utilize $\psi = 0.2$ and $\sigma = 0.05$ for all experiments. We include gray regions of uncertain cost as specified in the main text. For policy optimization, we implement PG-BROIL on top of the PPO implementation from~\cite{spinning-up} with all parameters set to their default settings except for $\alpha = 0.96$, policy learning rate set to  3e-4, and epochs set to 50.

\subsection{Reacher}
\label{app:reacher_details}
We build on the Reacher implementation from the DeepMind Control Suite~\cite{tassa2020dmcontrol} by adding a region with uncertain cost as specified in the main text.
For policy optimization, we implement PG-BROIL on top of the PPO implementation from~\cite{spinning-up} with all parameters set to their default settings except for $\alpha = 0.9$, policy learning rate set to 1e-4, hidden units set to 128, and epochs set to 800. To obtain preferences over the demonstrations, we rank each demonstration by the ground truth reward and assign pairwise preferences between each adjacent pair. Demonstrations were obtained by training a Soft Actor-Critic agent \cite{haarnoja2018soft} for 100 episodes and check-pointing the policy at each episode during training. This gives 100 demonstrations, and of these six with sufficiently different rewards were sampled.

\begin{figure}
     \centering
     \begin{subfigure}[b]{0.15\linewidth}
         \centering
         \includegraphics[width=\linewidth]{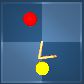}
         \caption{}
         \label{fig:demo_reacher}
     \end{subfigure}
     \hspace{2cm}
     \begin{subfigure}[b]{0.15\linewidth}
        \centering
         \includegraphics[width=\linewidth]{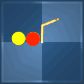}
         \caption{}
         \label{fig:training_reacher}
     \end{subfigure}
        \caption{Reacher environment during demonstration time (a) and policy training time (b). During demonstrations, the uncertain region (red) is far from the robot arm and the goal (yellow), but during policy optimization the goal position is randomized and sometimes the uncertain cost region is in the way forcing the agent to either go around or through it.} 
        \label{reacher_env}
\end{figure}

\subsection{TrashBot} \label{app:trashbot_details}
The TrashBot dynamics and actions are the same as in the Pointmass Navigation environment except that the system dynamics are deterministic. For policy optimization, we implement PG-BROIL on top of the PPO implementation from~\cite{spinning-up} with all parameters set to their default settings except for $\alpha = 0.95$, policy learning rate set to  3e-4, and epochs set to 50. 

\subsection{Atari Boxing} \label{app:boxing_details}
The Atari Boxing hyperparameters are the same as described in \ref{app:hyperparams} with $\alpha = 0.9$ and $\lambda = 0.3$ for PG-BROIL. We use a PG-BROIL implementation on top of the PPO implementation from \cite{spinning-up} with the default hyperparameters and epochs set to 800. To obtain preferences over the demonstrations, we rank each demonstration by its game score and assign pairwise preferences between each adjacent pair. Demonstrations were obtained by training a PPO agent with the standard hyperparameters in Table \ref{app:hyperparams} for 5 epochs and then taking four rollouts of episodes from the model.
%The pairwise preferences and demonstrations are as follows:
%\begin{figure}[h]
 %   \centering
  %  \includegraphics[width=0.5\linewidth]{figs/trash_figs/demonstrations.png}
   % \caption{Three pairs of preferences over ambiguous demonstrations}
    %\label{fig:trash_demos}
%\end{figure}
%\vskip -0.2in

\section{Baseline Algorithm Details}
\label{app:alg_details}
\paragraph{PBRL}\label{app:pbrl_details}
We implement PBRL by using the pairwise preference learning loss considered in~\cite{christiano2017deep}. We consider learning from offline preferences and build on the implementation from~\cite{brown2019extrapolating}. MCMC was performed for 20,000 steps with a proposal step size of 0.5. Weights are normalized so that $\Vert{w}\Vert_{1}=1$.

\paragraph{Bayesian REX}\label{app:brex_details}
We utilize the Bayesian REX implementation from~\cite{brown2020bayesian} to learn a Bayesian posterior over reward functions from offline preferences. MCMC was also performed for 20,000 sample steps with a proposal step size of 0.5. Weights are normalized so that $\Vert{w}\Vert_{1}=1$. We utilize a burn-in of 500 sample steps and down-sample to 20 samples.

\paragraph{GAIL}\label{app:gail_details}
We utilize the GAIL implementation from~\cite{GAIL_github}. We utilize PPO for policy optimization and use most of the default parameters from the provided implementation in~\cite{GAIL_github}. The only default parameters we changed were the L2 regularization coefficient for the weights of the discriminator network (set to $1e-2$), log std for the policy (set to $-0.5$), the hidden units of the policy network (set to $64$), and the total number of environment steps which we varied through a combination of changing the number of steps between each discriminator/policy update and the number of total iterations. We varied the number of environment steps and noted the behavior in  Table \ref{steps-table}. We found that on TrashBot with orders of magnitude more environmental steps we could not get consistently better performance across both trash collected and steps in the gray region so we report the performance with an equivalent number of environmental steps to PG-BROIL for all experiments.

\begin{table}
\caption{We run GAIL with differing number of environment steps and then compare PG-BROIL with GAIL with the same number of steps. Table \ref{sample-table} contains both GAIL and PG-BROIL with $2 \times 10^5$ steps. Results are averages ($\pm$ one st. dev.) over 100 test episodes each with a horizon of 100 steps per episode.}
\label{steps-table}
\vskip 0.15in
\begin{center}
\begin{small}
\begin{sc}
\scalebox{0.9}{\begin{tabular}{llll}
\toprule
algorithm & \begin{tabular}[c]{@{}l@{}}Number of Environment\\ Steps (x$10^5$) \end{tabular} & \begin{tabular}[c]{@{}l@{}}Avg. Trash\\ Collected\end{tabular} & \begin{tabular}[c]{@{}l@{}}Avg. Steps in \\ Gray Region\end{tabular} \\
\midrule
GAIL & 164 & 3.32 $\pm$ 1.66 & 0.35  $\pm$ 1.79 \\
GAIL & 41 & 2.88  $\pm$ 1.66 & 3.73  $\pm$ 7.98 \\
GAIL & \textbf{2} & 2.27  $\pm$ 1.66 & 5.08   $\pm$ 13.01 \\

PG-BROIL  & \textbf{2} & 9.20  $\pm$ 2.19 & 2.04  $\pm$ 5.94 \\
\bottomrule
\end{tabular}}
\end{sc}
\end{small}
\end{center}
\vskip -0.1in
\end{table}

\paragraph{BC}\label{app:bc_details}
We utilize the same stochastic policy and learning rate scheduler as for PPO but simply maximize the log-likelihood of actions in each of the states in the demonstrations. The learning rate for the policy is $1e-2$ and the number of BC iterations is $1000$.

\section{TrashBot Further Analysis and Visualization}\label{app:trashbot_analysis}
\subsubsection{Example Rollouts}\label{app:rollouts}
\begin{figure}[!ht]
\centering
\includegraphics[width=.32\textwidth]{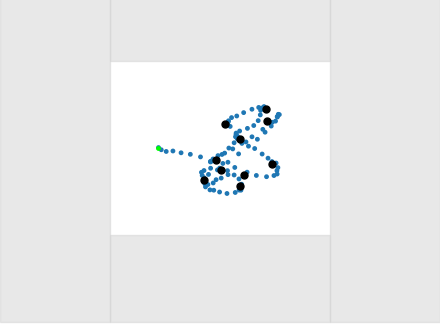}
\includegraphics[width=.32\textwidth]{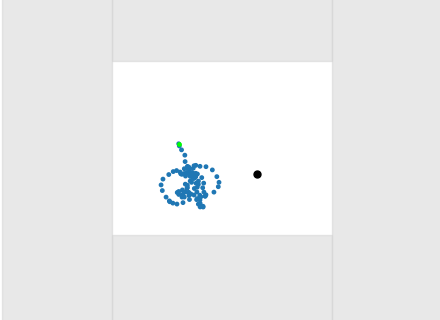}
\includegraphics[width=.32\textwidth]{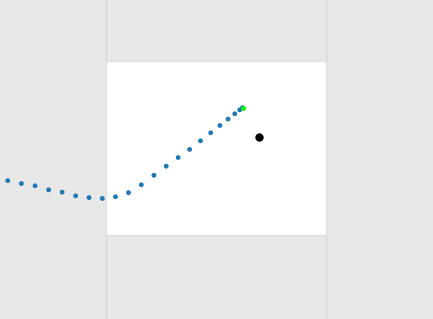}
\caption{\textbf{PG-BROIL: }The left and middle images show trajectories for PG-BROIL with lambda value of $0.8$ while the right image shows a failure case for lambda value of $0.7$. PG-BROIL is able to successfully pick up trash and avoid excessive steps in the gray region by hedging against all reward hypotheses with sufficient probability, allowing it to recognize that it is more important to collect trash than simply stay in the white region.}
\label{fig:pgbroil-rollouts}
\end{figure}

\begin{figure}[!ht]
\centering
\includegraphics[width=.32\textwidth]{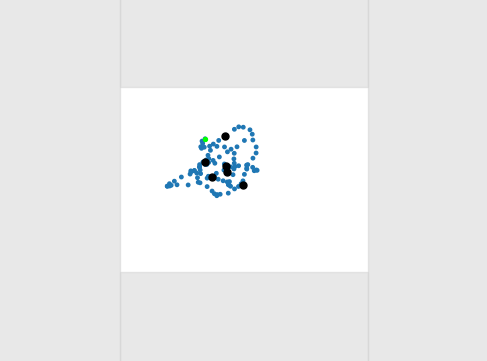}
\includegraphics[width=.32\textwidth]{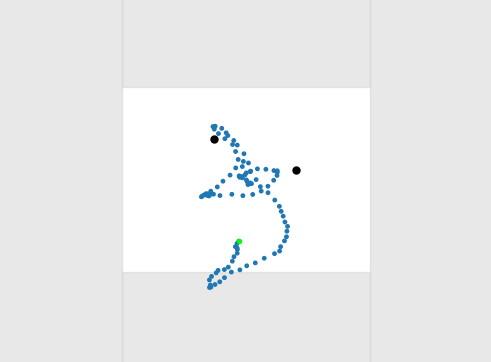}
\includegraphics[width=.32\textwidth]{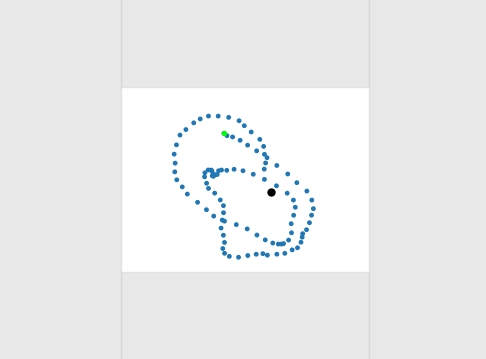}
\caption{\textbf{PBRL: }PBRL places a small positive weight on staying in the white region, resulting in it falling in a local minima where it mostly optimizes for staying in the white region rather than collecting trash. This leads to low visitation of the gray region as desired, but relatively inconsistent performance in picking up pieces of trash.}
\label{fig:pbrl-rollouts}
\end{figure}

\begin{figure}[!ht]
\centering
\includegraphics[width=.32\textwidth]{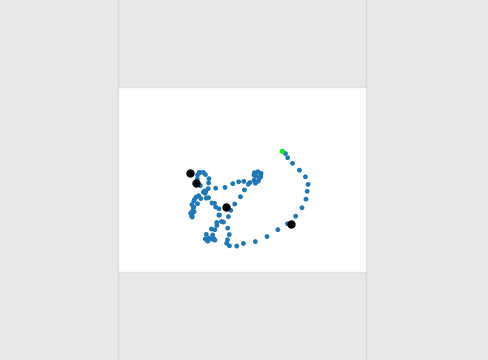}
\includegraphics[width=.32\textwidth]{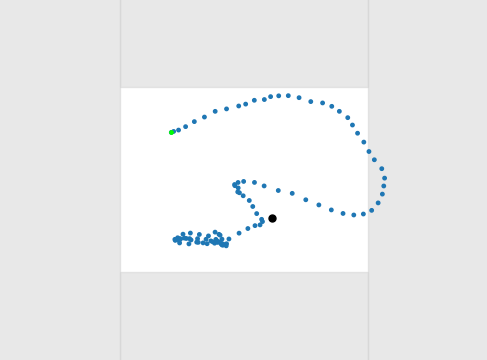}
\includegraphics[width=.32\textwidth]{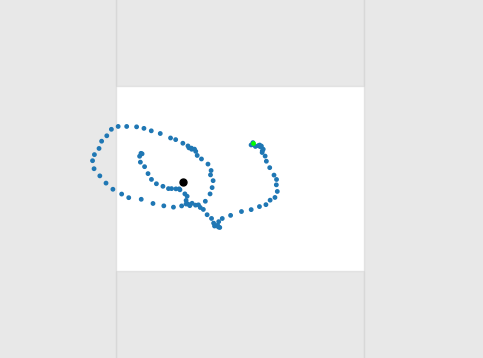}
\caption{\textbf{Bayesian REX: }Bayesian REX weights picking up trash and staying in the white region roughly equally. Thus, Bayesian REX explores the entire white region, not just the central portion where the trash is located, resulting in frequent forays into the gray region.}
\label{fig:brex-rollouts}
\end{figure}

\begin{figure}[!ht]
\centering
\includegraphics[width=.32\textwidth]{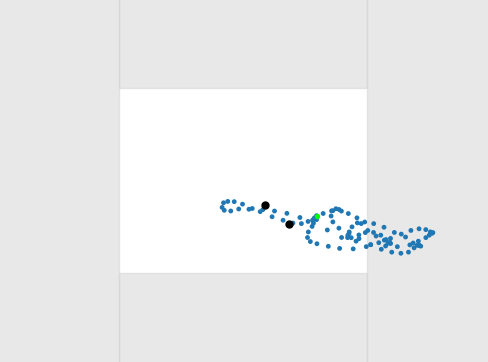}
\includegraphics[width=.32\textwidth]{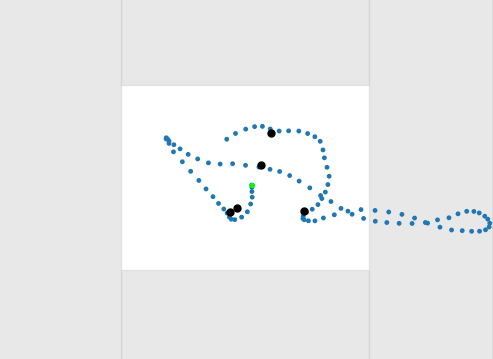}
\includegraphics[width=.32\textwidth]{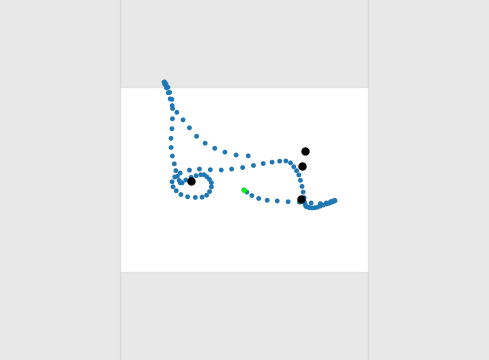}
\caption{\textbf{GAIL: } The left, middle and right images show an example trajectory from GAIL running with $2 \times 10^5$, $4.1 \times 10^6$, and $1.64 \times 10^7$ environment steps respectively. Due to the lack of environment steps, the bot in the left and middle images take more steps in the gray region before turning around and going back to the white region. However, the bot in right image immediately turns around as soon as it contacts the gray region. The bot in the middle and right images also collect more trash in their episodes than the left image. This behavior is consistent with the averages in Table \ref{steps-table}.}
\label{fig:gail-rollouts}
\end{figure}

\begin{figure}[!ht]
\centering
\includegraphics[width=.32\textwidth]{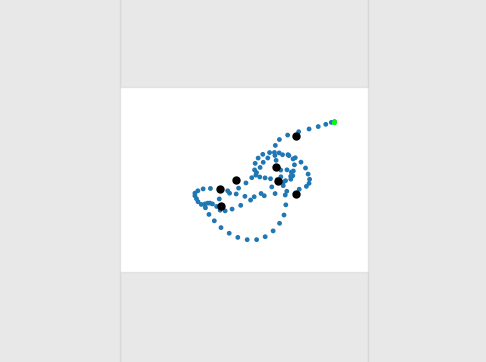}
\includegraphics[width=.32\textwidth]{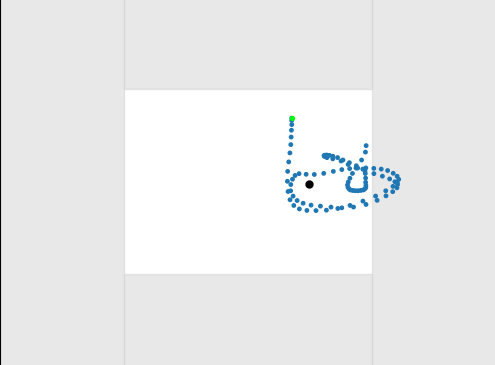}
\includegraphics[width=.32\textwidth]{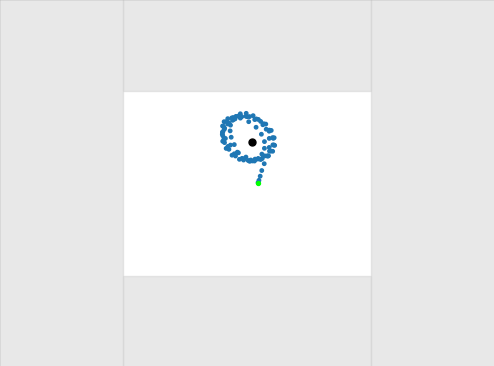}
\caption{\textbf{BC: }The failure cases come from one of the demonstrations having the same behavior of circling the trash without picking it up. Since BC is only incentivized to exactly mimic the actions in demonstration states, it is unable to navigate ambiguities in the demos.}
\label{fig:bc-rollouts}
\end{figure}

In Figures~\ref{fig:pgbroil-rollouts}-\ref{fig:bc-rollouts} we show both successful and unsuccessful rollouts from fully trained policies for PG-BROIL and all baselines to gain intuition for their quantitative performance. In all rollouts below, on the left we show a successful case while the middle and right images are failure cases. As noted in the experiments section in the main text, PBRL places a small positive weight on staying in the white region, resulting in it falling in a local minima where it mostly optimizes for staying in the white region rather than collecting trash. This leads to low visitation of the gray region as desired, but relatively inconsistent performance in picking up pieces of trash. Bayesian REX on the other hand weights picking up trash and staying in the white region roughly equally. Thus, Bayesian REX explores the entire white region, not just the central portion where the trash is located, resulting in frequent forays into the gray region. PG-BROIL is able to successfully pick up trash and avoid excessive steps in the gray region by hedging against all reward hypotheses with sufficient probability, allowing it to recognize that it is more important to collect trash than simply stay in the white region. 

\vspace{1cm}

\subsubsection{Posterior Analysis}\label{app:posterior_analysis}
\begin{figure}
    \centering
    \includegraphics[width=0.9\linewidth]{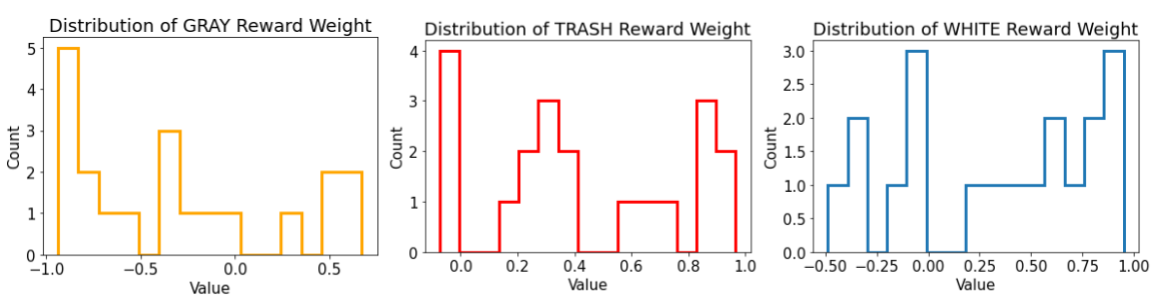}
    \caption{Distribution of each feature weight in posterior for seed 0.}
    \label{fig:weights_hist}
\end{figure}

Figure \ref{fig:weights_hist} shows the distribution of the weights for each feature for PG-BROIL. PG-BROIL exploits the fact that some reward functions have a negative weight for the WHITE feature to recognize that simply staying in the white region without going for trash is a highly suboptimal strategy. This allows PG-BROIL to outperform PBRL, which falls into a local maxima by simply mining rewards by staying in the white region.

Additionally, amongst the 20 reward functions generated on seed 0, the WHITE and TRASH features have a Pearson correlation coefficient of -0.46. This implies that if a reward function places high weight on the WHITE feature, it is likely to place a smaller or more negative weight on the TRASH feature and vice-versa. This helps create the causal confusion we see in this experiment since it is unclear whether the agent should be rewarded more for the WHITE feature or the TRASH feature.

\subsubsection{Sensitivity to $\lambda$}\label{app:lambdas}

\begin{figure}
\centering
\includegraphics[width=.45\linewidth]{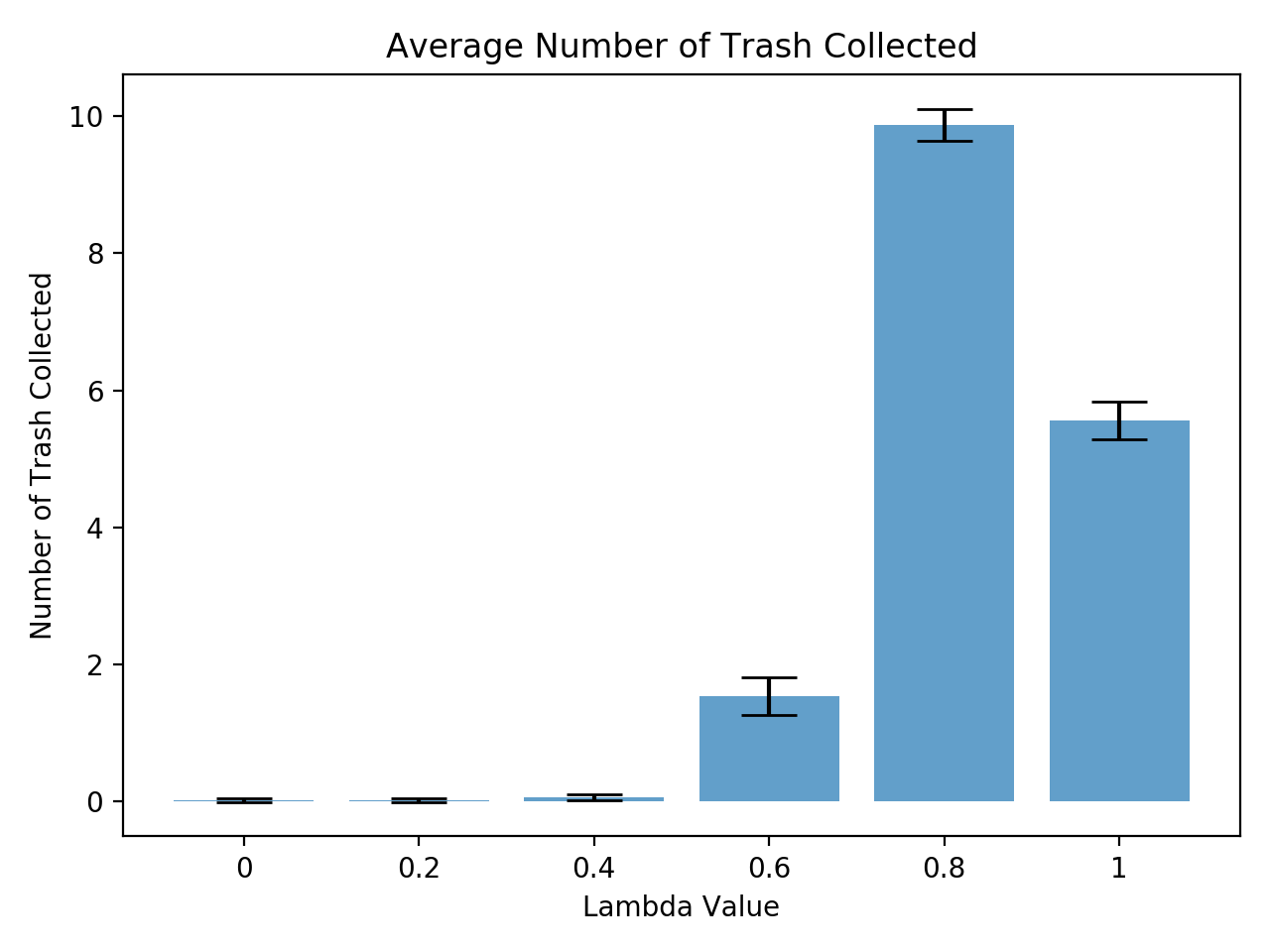}
\includegraphics[width=.45\linewidth]{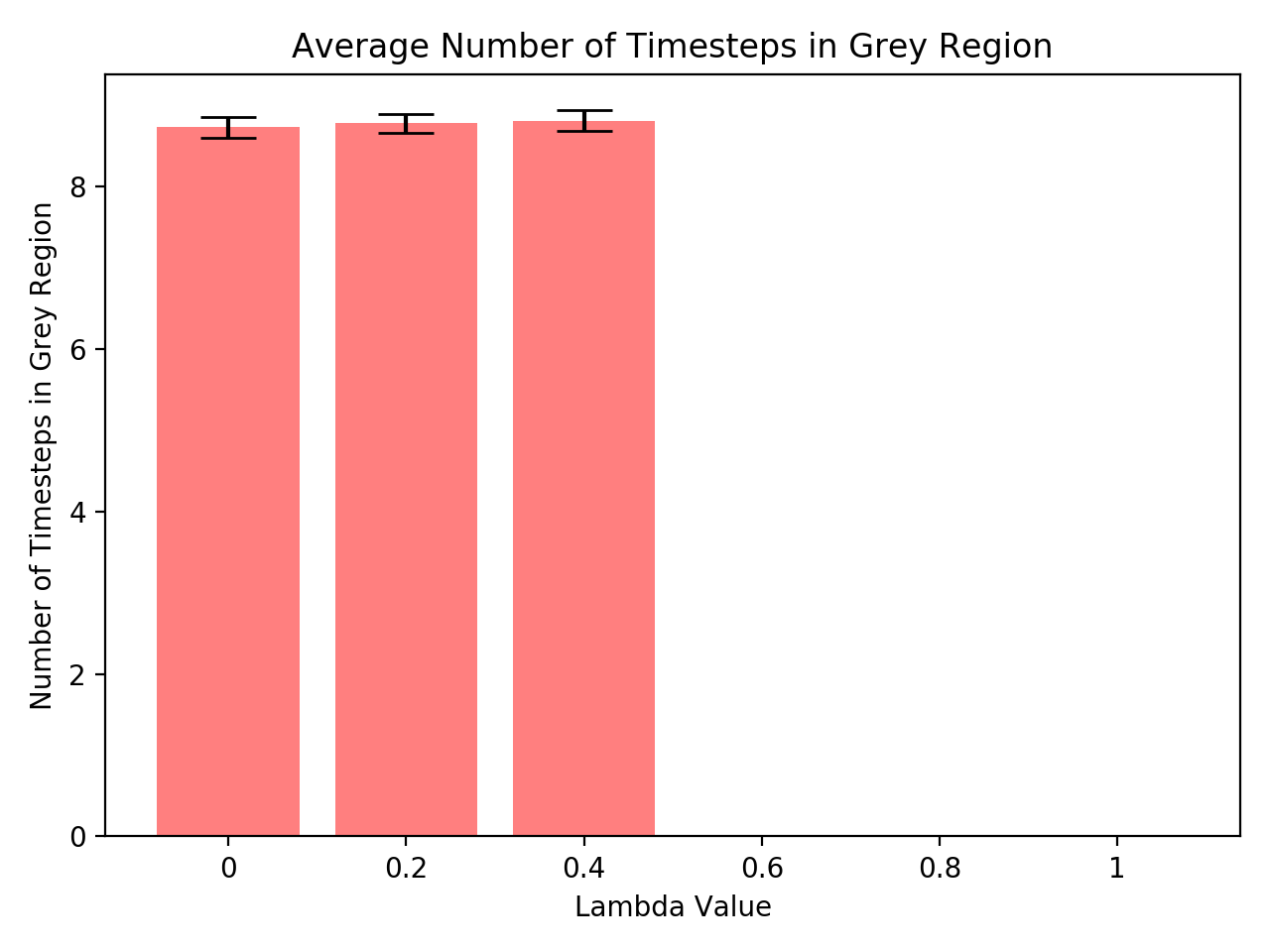}
\caption{We run the TrashBot environment with CVaR risk metric and $\alpha = 0.95$ over various $\lambda$. We take the 95\% confidence interval and plot them as the error bars. We find that the TrashBot collects the most trash with the minimum amount of timesteps spent in the grey region when $\lambda = 0.8$.}
\label{fig:trashbot_lambdas}
\end{figure}

Figure \ref{fig:trashbot_lambdas} shows the TrashBot experiment results over various values of $\lambda$. We found $\lambda=0.8$ to give the best performance in terms of trash collection and gray space avoidance.

\section{Sensitivity to Alpha}
Most applications of CVaR use $\alpha \in [0.9,1)$ since as $\alpha \rightarrow 0$ CVaR is equivalent to expected value. Empirically, we found that $\alpha>0.8$ is required to get behaviors different from those that simply maximize expected reward, i.e., $\lambda$ has little to no effect on the resulting policy behavior for $\alpha \leq 0.8$. 
\end{document}